%% file: deoct.tex
\documentclass{article}

\usepackage[preprint]{neurips_2023}

\usepackage[utf8]{inputenc} 
\usepackage[T1]{fontenc}    
\usepackage{hyperref}       
\usepackage{url}            
\usepackage{booktabs}       
\usepackage{amsfonts}       
\usepackage{nicefrac}       
\usepackage{microtype}      
\usepackage[dvipsnames]{xcolor}         
\definecolor{mygreen}{RGB}{56, 173, 107}

\usepackage{amsmath}
\usepackage{amssymb}
\usepackage{mathtools}
\usepackage{amsthm}
\usepackage{multirow}
\usepackage{setspace}
\theoremstyle{plain}

\theoremstyle{definition}

\theoremstyle{remark}

\usepackage[capitalize,noabbrev]{cleveref}
\usepackage{algorithm}
\usepackage{algorithmic}
\usepackage[flushleft]{threeparttable}
\usepackage{multicol}
\usepackage{tikz}
\usetikzlibrary{decorations.pathreplacing,calc}

\usepackage{natbib}
\setcitestyle{authoryear, open={(},close={)}}

\usepackage{longtable}
\usepackage{chngpage}

\title{A GPU-Accelerated Moving-Horizon Algorithm for Training Deep Classification Trees on Large Datasets}

\author{%
    Jiayang Ren$^{\dag}$, Valentín Osuna-Enciso$^{\ddag}$, Morimasa Okamoto$^{\dag}$, Qiangqiang Mao$^{\dag}$, \\
    \textbf{Chaojie Ji}$^{\dag}$, \textbf{Liang Cao}$^{\dag}$, \textbf{Kaixun Hua}$^{\dag}$, \textbf{Yankai Cao}$^{\dag}$\thanks{corresponding author: yankai.cao@ubc.ca.} \\
      $^{\dag}$University of British Columbia\\
      $^{\ddag}$Universidad de Guadalajara
}

\begin{document}

\maketitle

\begin{abstract}
Decision trees are essential yet NP-complete to train, prompting the widespread use of heuristic methods such as \texttt{CART}, which suffers from sub-optimal performance due to its greedy nature. Recently, breakthroughs in finding optimal decision trees have emerged; however, these methods still face significant computational costs and struggle with continuous features in large-scale datasets and deep trees. To address these limitations, we introduce a moving-horizon differential evolution algorithm for classification trees with continuous features (\texttt{MH-DEOCT}). Our approach consists of a discrete tree decoding method that eliminates duplicated searches between adjacent samples, a GPU-accelerated implementation that significantly reduces running time, and a moving-horizon strategy that iteratively trains shallow subtrees at each node to balance the vision and optimizer capability. Comprehensive studies on 68 UCI datasets demonstrate that our approach outperforms the heuristic method \texttt{CART} on training and testing accuracy by an average of 3.44\% and 1.71\%, respectively. Moreover, these numerical studies empirically demonstrate that \texttt{MH-DEOCT} achieves near-optimal performance (only 0.38\% and 0.06\% worse than the global optimal method on training and testing,  respectively), while it offers remarkable scalability for deep trees (e.g., depth=8) and large-scale datasets (e.g., ten million samples).

\end{abstract}

\section{Introduction} 
\label{sec:intro}
Decision trees, as a supervised learning technique, offer exceptional interpretability, making them suitable for many domains requiring clear decision-making \citep{morgan1963}. Their tree-like structure systematically partitions the feature space to address classification or regression problems. However, building an optimal decision tree is an NP-complete problem \citep{laurent1976}, leading to heuristic strategies like \texttt{ID3} \citep{quinlan1986}, \texttt{C4.5} \citep{quinlan1993}, and \texttt{CART} \citep{breiman1984}. These methods utilize a top-down, greedy approach to construct trees. Despite their widespread use, heuristic techniques are limited by their focus on individual tree splits, neglecting potential impacts of future splits \citep{bertsimasOptimal2017}. Consequently, these algorithms risk arriving at sub-optimal solutions, potentially compromising the decision tree's predictive capabilities.

Addressing greedy problems, optimal classification trees (OCT) that induce the entire tree at once have gained significant attention in recent years. Two primary approaches to solving OCT are mixed integer programming (MIP) \citep{bertsimasOptimal2017, verwer2017, verwer2019a, gunluk2021, aghaei2022, hua2022a, mazumder2022} and dynamic programming (DP) \citep{hu2019b, aglin2020a, lin2020c, demirovic2022a, mctavish2022a}. MIP methods learn optimal trees with a fixed depth as a MIP model and employ general or specific-purpose MIP solvers to attain global optimum. State-of-the-art methods \texttt{RS-OCT} \citep{hua2022a} and \texttt{Quant-BnB} \citep{mazumder2022} expanded solvable regions to datasets with one million samples through well-designed lower and upper bounds and sophisticated branch-and-bound frameworks. However, these methods still face substantial computational costs, as they can only handle shallow decision trees with two or three layers. \texttt{RS-OCT} offers theoretical convergence to the global optimal solution but relies heavily on the parallelization of large computing clusters (e.g., 1000 cores). \texttt{Quant-BnB} exhibits minimal computing time for two-layer trees but is limited to shallow trees up to three layers, due to the exponential increase in both tree nodes and computing time with each increment in tree depth. As for DP, studies conducted have demonstrated promising results in solving OCT problems. Nonetheless, the majority of these DP approaches can only deal with shallow trees and primarily focus on binary features \citep{demirovic2022a}. Converting continuous features into binary features can lead to a significantly expanded feature space, potentially disrupting the relationship between values of numerical features \citep{mazumder2022}. In conclusion, existing OCT methods face difficulties when computing optimal decision trees with continuous features, particularly for large-scale datasets and deep trees. 

In addition to OCT that aim to achieve the optimal solution, there are alternative approaches focused on enhancing greedy algorithms by introducing stochastic elements into the process. One noteworthy method in this category is the local search method proposed by \citep{bertsimasOptimal2017}. Similar to the conventional approach of \texttt{CART}, the local search method traverses a single current node to generate an initial solution. However, it diverges in its strategy for refinement—improving the initial tree through node swapping and continuous evaluation of the objective function until further enhancement is unattainable. Empirical studies conducted on extensive datasets demonstrate that local search consistently outperforms \texttt{CART}. Nevertheless, the computational overhead introduced by node swapping renders local search less efficient, especially when dealing with large datasets and deep trees. Additionally, the implementation's reliance on a caching technique complicates parallelization on modern computers. A more detailed overview of related works can be found in Appendix A.

In this paper, we introduce a gpu-accelerated moving-horizon differential evolution algorithm for optimal univariate binary trees on classification tasks with continuous features (\texttt{MH-DEOCT}). Differential evolution (DE) is a metaheuristic method that can search extensive spaces of candidate solutions for a black box optimization problem. There are several metaheuristic-based methods proposed for inducing decision trees \citep{kuo2007, cho2011a, dolotov2020, ersoy2020, rivera-lopez2022}. However, these methodologies require searching the entire tree simultaneously, while the tree nodes increases exponentially with tree depth. Given that black-box optimization methods are typically designed to handle a limited number of variables, these methodologies tend to underperform when applied to deep trees. In our experiments, the plain DE algorithm, without the moving-horizon strategy and warm-starts, could not even outperform \texttt{CART} when the depth was 8. To address this issue, we propose utilizing a moving-horizon strategy that optimizes shallow subtrees at a time to balance the vision of trees and the ability of optimizers. 

\textbf{Our contributions:} First, we devise a discrete tree decoding method from DE individuals that eliminates duplicated searches between adjacent samples, thereby enhancing the efficiency in handling continuous features. Second, we implement a population-based, GPU-accelerated fitness evaluation algorithm for OCT along with the corresponding population-based DE algorithm. This powerful combination drastically reduces the running time. Finally, we propose a moving-horizon strategy that optimizes shallow subtrees at each node incrementally to strike a balance between the vision and the optimizer capability. 

\textbf{Performance:} Our comprehensive studies on 68 UCI datasets demonstrate that the proposed \texttt{MH-DEOCT} outperforms heuristic methods like \texttt{CART}, enhancing training and testing accuracy by an average of 3.44\% and 1.71\%, respectively. Moreover, these numerical studies empirically demonstrate that \texttt{MH-DEOCT} achieves near-optimal performance with significantly reduced running time, trailing the global method \texttt{Quant-BnB} by an average of 0.38\% and 0.06\% on training and testing, respectively. Furthermore, our approach extends its performance to deeper trees (e.g., depth=8) and large-scale datasets (e.g., ten million samples), making it a promising solution for a broad range of applications.

\section{Preliminaries and Notations}
\textbf{Differential Evolution} \label{sec:de} is a population-based meta-heuristics optimization algorithm consisting of four operators, including initialization, mutation, crossover, and selection \citep{Storn_1997_Differential}. For an optimization problem with a $m$-dimensional solution, DE starts by randomly initializing a population consisting of $N$ vectors with $m$ dimension corresponding to the solution. Each vector in this population is called an individual. We denote the population at the $g$th generation as $\mathcal{S}^{g} = \{\mathbf{s}_1^g, \mathbf{s}_2^g,\cdots, \mathbf{s}_N^g \}$ and the $r$th individual at the $g$th generation as $\mathbf{s}_r^g = [s_{r,1}^g, s_{r,2}^g,\cdots, s_{r,m}^g] \in \mathbb{R}^m$. After population initialization, the mutation operator generates mutant vectors $\mathbf{v}_r^g = [v_{r,1}^g, v_{r,2}^g,\cdots, v_{r,m}^g] \in \mathbb{R}^m$ from the parent population $\mathcal{S}^{g}$:
\begin{equation}\label{eqn:mutation}
	\mathbf{v}_r^g = \mathbf{s}_{best}^{g} + M\cdot (\mathbf{s}_{r_1}^g-\mathbf{s}_{r_2}^g)
\end{equation}
where $\mathbf{s}_{best}^g$ is the individual with the best fitness in the current generation, $r_1, r_2 \in \{1,\cdots,N\}$ are two randomly selected indices and $M$ is the mutation factor. Subsequently, an offspring individual $\mathbf{u}_r^{g} = [u_{r,1}^{g}, u_{r,2}^{g}, \cdots, u_{r,m}^{g}] \in \mathbb{R}^m$ is generated by the crossover operation:
\begin{equation}\label{eqn:crossover}
	u_{r,q}^{g} =
	\begin{cases}
		v_{r,q}^g & \text{, if $rand(0,1) \leq CR_r$ \ or\ $q = q_{rand}$ }\\
		s_{r,q}^g & \text{, otherwise}
	\end{cases} 
\end{equation}
where $rand(0,1)$ is a uniform random number within a range of [0,1] for each individual $r\in\{1, \cdots, N\}$ and each dimension $q\in\{1,\cdots, m\}$, $CR_r$ is a pre-specified cross probability, $q_{rand}$ is a random index chosen from $\{1, \cdots ,m \}$. Then, a fitness value $f(\mathbf{s}_r^g)$ is used to perform the selection: 
\begin{equation}\label{eqn:selection}
	\mathbf{s}_{r}^{g+1} =
	\begin{cases}
		\mathbf{u}_r^g & \text{, if $f(\mathbf{u}_r^g) \leq f(\mathbf{s}_r^g)$ } \\ 
		\mathbf{s}_r^g & \text{, otherwise}
	\end{cases}       
\end{equation}
The above procedure iterates until the maximum generation number $G$ arrives. A detailed description of the classic DE procedure can be found in Appendix \ref{apx:classicde}, \cref{alg:classicde}.

\textbf{Optimal Classification Tree (OCT): } \label{sec:oct} 
We consider univariate binary trees with continuous features and depth $D$ for classification. "Binary" implies each node has at most two child nodes. "Continuous" differs from "categorical" features in the dataset, while the later one can be incorporated using encoding methods like one-hot encoding. "Univariate" denotes that only one feature can be selected for splitting at each node. 

To facilitate ease of reading, we primarily adopt the notation system used in \cite{bertsimasOptimal2017}. Denote the training dataset as $(\mathbf{X}, \mathbf{Y})$ and the samples in the dataset as $(\mathbf{x}_i, y_i)$, where $i=1,\cdots,n$ and $n$ is the number of samples. Each sample consists of a pair of $P$ features $\mathbf{x}_i \in \mathbb{R}^P$ and a class indicator $y_i \in \mathcal{K} = \{1, \cdots, K\}$, with $K$ representing the total number of classes. Notably, we scale $\mathbf{x}_i$ with the min-max value of each feature, resulting in a range of $[0,1]$.

\cref{fig:dt_chr} illustrates a typical example of a binary tree with depth $D=3$. The tree comprises branch nodes that apply a split and leaf nodes that assign a class. The total number of nodes in a complete tree with depth $D$ is $T=2^{(D+1)}-1$. Each node in the binary tree is assigned an index $t\in\mathcal{T}=\{1,\cdots,T\}$, counted from top to bottom and left to right, as shown in \cref{fig:dt_chr}. We further denote the index set of branch nodes as $\mathcal{T}_B = \{1,\cdots,\lfloor T/2 \rfloor\}$ and leaf nodes as $\mathcal{T}_L = \{\lfloor T/2 \rfloor + 1,\cdots,T\}$. 

The univariate binary tree is described using a vector tuple $(\mathbf{a}, \mathbf{b})$, where $\mathbf{a}=[a_1,\cdots,a_{\lfloor T/2 \rfloor}]$ and $\mathbf{b}=[b_1,\cdots,b_{\lfloor T/2 \rfloor}]$. Specifically, $a_t\in \{0,1,\cdots,P\}, t\in \mathcal{T}_B$, represents the selected feature to split at branch node $t$. If $a_t=0$, it means there is no split at this node. $b_t\in [0,1], t\in \mathcal{T}_B$ represents the threshold to split at branch node $t$. For a branch node $t$ and a sample $\mathbf{x}_i$, if the $a_t$th feature $x_{i, a_t}<b_t$, the sample is assigned to the left child node; otherwise, it is assigned to the right child node. To maintain the complete structure of a decision tree with depth $D$, we assign all samples at branch node $t$ to the right child node if no feature is selected by forcing $b_t=0$ when $a_t=0$.

We utilize $z_{i,t} = 1$ to indicate that an sample $\mathbf{x}_i$ is assigned to a leaf node $t\in \mathcal{T}_L$ following the assignment procedure described above; otherwise, $z_{i,t} = 0$. Besides, $z_{i,t}^k$ is denoted as the class flag representing if sample $\mathbf{x}_i$ of class $k$ is assigned to the leaf node $t$, with $z_{i,t}^k=1$ if $z_{i,t} = 1$ and $y_i=k$, otherwise, $0$. In this manner, the assigned class of each leaf node $c_t\in \{1,\cdots,K\}, t\in \mathcal{T}_L$ can be calculated using:
\begin{equation} \label{eqn:opt_c}
 c_t = \arg \max_k \{\hat{z}_t^k\},\; \hat{z}_t^k=\sum_{i=1}^n z_{i,t}^k, \; k\in\mathcal{K}, t \in \mathcal{T}_L,
\end{equation}
where $\hat{z}_t^k$ is the number of assigned samples that belong to class $k$ on the leaf node $t$. The optimal classification tree aims to find the optimal solution to the following problem:
\begin{subequations}\label{eqn:opt_all}
	\begin{align}
        \min_{\mathbf{a}, \mathbf{b}} & \;\; \sum_{i=1}^n \ell_{0-1}(\hat{y}_i, y_i) + \alpha\sum_{t\in\mathcal{T}_B} \mathbf{1}_{\mathbb{N}}(a_t), \label{eqn:opt_loss} \\ 
        & \text{s.t.} \;\; \hat{y}_i = f_{tree}(\mathbf{a}, \mathbf{b}, \mathbf{x}_i) \\
        & \;\;\;\;\;\; n_t\geq n_{min}\times \mathbf{1}_{\mathbb{N}}(n_t), t\in \mathcal{T}_L \label{eqn:opt_nmin}
    \end{align}
\end{subequations}
where $\hat{y}_i = f_{tree}(\mathbf{a}, \mathbf{b}, \mathbf{x}_i)$ is the predicted class of sample $\mathbf{x}_i$ through the univariate binary tree $(\mathbf{a}, \mathbf{b})$; $f_{tree}(\mathbf{a}, \mathbf{b}, \mathbf{x}_i)=c_t$ for $t$ with $z_{i,t}=1$; $\ell_{0-1}$ is a 0-1 misclassification loss, where $\ell_{0-1}(\hat{y}_i, y_i)=1$ if $\hat{y}_i \neq y_i$, and $0$ otherwise; $\mathbf{1}_{\mathbb{N}}(\cdot)$ is the indicator function on the set of natural numbers; $\sum_{t\in\mathcal{T}_B} \mathbf{1}_{\mathbb{N}}(a_t)$ is the complexity of the tree representing the number of splits included in the tree; $\alpha$ is the complexity penalty parameter;  $n_t=\sum_{i=1}^n z_{i,t}$ is the number of samples assigned to leaf node $t$, and constraint \ref{eqn:opt_nmin} ensures $n_t$ is larger than the minimum sample size $n_{min}$ when $n_t\neq 0$. From the perspective of leaf nodes, the misclassification loss in the cost \ref{eqn:opt_loss} can also be represented as: $\sum\limits_{i=1}^n \ell_{0-1}(\hat{y}_i, y_i) =  n - \sum\limits_{t\in\mathcal{T}_L} \max\limits_{k\in\mathcal{K}}{\hat{z}_t^k}$.

\section{Differential Evolution for Optimal Classification Tree (DEOCT)}

\subsection{Decoding Tree Candidates from Individuals Discretely}
In differential evolution, the individual $\mathbf{s}_r$ is a real-valued vector. However, the univariate binary tree consists of a set of nodes with an integer scalar $a_t$, a real-valued scalar $b_t$. Therefore, we need to decode the univariate binary trees from DE individuals. Here, the individual of a univariate binary tree with the depth of $D$ is consist of $\mathbf{s}_r = [\hat{a}_{1},\cdots,\hat{a}_{t},\cdots,\hat{a}_{\lfloor T/2 \rfloor},\hat{b}_1,\cdots,\hat{b}_t,\cdots,\hat{b}_{\lfloor T/2 \rfloor}]$, where $\hat{a}_{t}\in [0, P+1)$, $\hat{b}_{t}\in [0,1), t\in \mathcal{T}_B$. 

Sequentially, we can decode the selected feature $a_t$ in a tree from $\hat{a}_t$ in a DE individual:
\begin{equation}\label{eqn:aenc}
	\begin{aligned}
		a_{t} & = \lfloor \hat{a}_{t} \rfloor, \;\; \hat{a}_t\in[0,P+1), \;\; t \in \mathcal{T}_B \\
	\end{aligned}
\end{equation}
As for the split threshold $b_t$, it is obviously that we can directly utilize $b_t = \hat{b}_t$ to decode. However, this direct method will introduce duplicated splits between two adjacent samples, which won't influence the training loss. Hence, instead of directly decoding the values of $b_t$, we propose a discrete decoding approach that utilize $\hat{b}_t$ to represent the index of $b_t$ in a sorted threshold set.  
\vskip -0.05in
\begin{figure}[!htp]
	\begin{center}
        \centerline{\includegraphics[width=0.95\columnwidth]{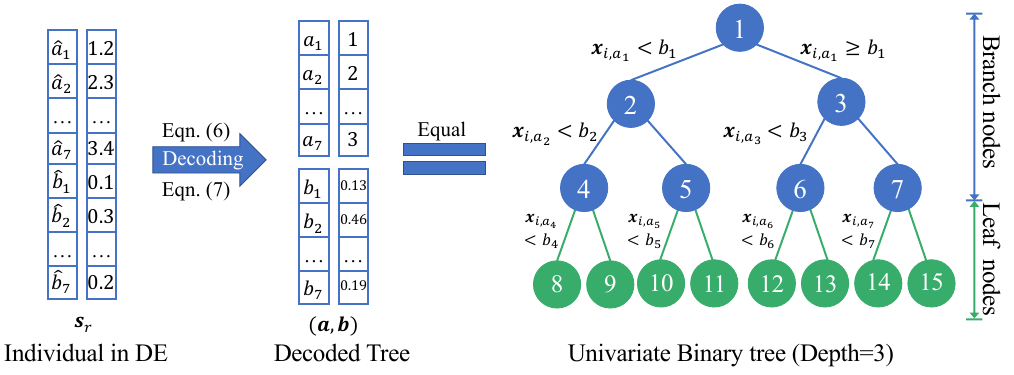}} 
		\caption{Example of a binary tree with depth=3 and the decoding procedure from a DE individual}
		\label{fig:dt_chr}
	\end{center}
  \vskip -0.2in
\end{figure}

Specifically, we denote $\hat{x}_{i, p}$ as the unique elements in the $p$th feature of the original dataset, subject to the constraints $\hat{x}_{i, p}<\hat{x}_{i+1, p}, i=1,\cdots,n_p$. Here, $n_p$ is the number of unique elements in the $p$th feature. Then, the threshold set for $p$th feature is generated by splitting at the mean values of adjacent $\hat{x}_{i, p}$, which is $\mathcal{B}_p := (\beta_{p,i})_{i=1}^{n_p+1} = (0, \frac{\hat{x}_{1, p} + \hat{x}_{2, p}}{2}, \cdots, \frac{\hat{x}_{n_{p-1}, p} + \hat{x}_{n_p, p}}{2}, 1)$. Finally, the split threshold $b_t$ for node $t$ with the split feature $a_t$ is decoded from $\hat{b}_t$ utilizing the following equation:

\begin{equation}\label{eqn:benc}
\begin{aligned}
    b_t = \beta_{a_t,i},\; i = \lfloor \hat{b}_t \times (n_p+1)\rfloor + 1, \; t\in \mathcal{T}_B,
    \end{aligned}
\end{equation}
where $a_t$ is the selected feature to split, $i$ is the index of split threshold $b_t$ in the threshold set $\mathcal{B}_p$. Note that these threshold sets only need to generate once at the beginning of the algorithm, ensuring minimal additional costs. \cref{fig:dt_chr} shows an example of a univariate binary tree $(\mathbf{a}, \mathbf{b})$ with depth $D=3$ decoded from an individual $\mathbf{s}_r$. 

\subsection{Parallel Fitness Evaluation of Tree Candidates Using GPU}\label{sec:fitness}
\vskip -0.05in
\begin{figure}[!htp]
	\begin{center}
    \centerline{\includegraphics[width=0.95\columnwidth]{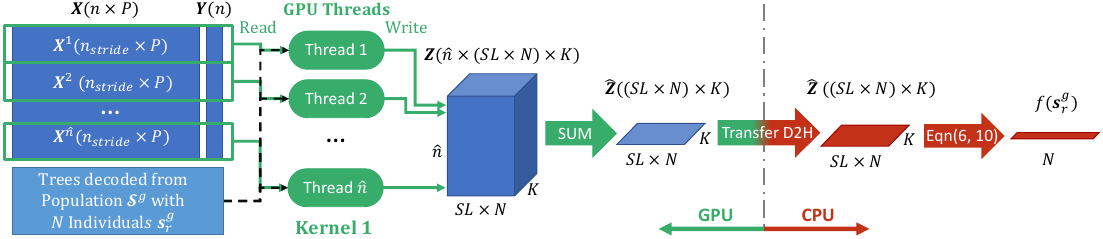}} 
		\caption{SIMD parallelization model for OCT fitness evaluation in GPU (Line 6-20 in \cref{alg:oct_fit})}
		\label{fig:gpu}
	\end{center}
  \vskip -0.15in
\end{figure}

Differential Evolution relies on individual fitness values $f(\mathbf{s}_r)$ to select the best candidate over generations. A key challenge in applying DE to the OCT problem is designing a suitable fitness function. Since DE does not support constraints other than range constraints, we transform constraints (\ref{eqn:opt_nmin}) in the OCT problem (\ref{eqn:opt_all}) into soft constraints within the objective function as $f(\mathbf{s}^g_r) =  n - \sum\limits_{t\in\mathcal{T}_L} \max\limits_{k\in\mathcal{K}}{\hat{c}_t^k} + \alpha\sum\limits_{t\in\mathcal{T}_B} \mathbf{1}(a_t) + \sum\limits_{t\in \mathcal{T}_L} V_{t}$, where $V_t=1$ if $ n_t < n_{min}\times \mathbf{1}_{\mathbb{N}}(n_t)$, otherwise, 0.  

The OCT fitness function's computational cost significantly impacts the overall training time, as DE requires fitness evaluation for each individual. Most heuristic methods like \texttt{CART} employ a sorted and cached approach to reduce costs. However, DE relies on random individual movements and cannot utilize this method for OCT fitness calculation. Fortunately, examining the fitness function $f(\mathbf{s}^g_r)$, we notice that operations for each sample $\mathbf{x}_i$ are identical, enabling GPU parallelization using SIMD (single instruction, multiple data). \cref{alg:oct_fit} and \cref{fig:gpu} detail the procedure for computing fitness values for a population $\mathcal{S}^g$. The algorithm starts by decoding trees from individuals, transferring necessary data to GPU's global memory, and processing the dataset in parallel. Then, the class count matrix $\hat{\mathbf{Z}}$, representing the number of assigned samples that belong to each class on the leaf nodes, is aggregated and transferred back to the host. Finally, the fitness values for individuals are calculated in CPU.

The algorithm's efficiency comes from leveraging the parallel architecture of GPUs. Notably, we process $n_{stride}$ adjacent samples in one thread, and store the class flags' sum value in $\mathbf{Z}$. Here, $\mathbf{Z}$ is the matrix of the class flag $z_{i,t}^k$ representing if sample $i$ of class $k$ is assigned to the leaf node $t$. This operation reduces expensive random access for adjacent samples. The dimension of $\mathbf{Z}$ is also reduced for faster sum operations in Line 17. Furthermore, by calculating fitness for the entire population at once, we only need to read the adjacent samples once for all the individuals.

\subsection{Optimal Classification Tree Induction via Differential Evolution}
As mentioned in \cref{sec:fitness}, we need to modify the classic differential evolution algorithm to treat the entire population as the fundamental unit for fitness evaluation in GPU. To achieve this, we first generate a new generation population by performing mutation and crossover operations based on the best individual from the previous generation, rather than the current best one as illustrated in Equation \ref{eqn:mutation}. We then employ the GPU-parallelized OCT fitness evaluation (Algorithm \ref{alg:oct_fit}) to assess the fitness of the new individuals and select the better individual accordingly. This approach generates a non-descending sequence of fitness values, ultimately leading to a near-optimal solution. Furthermore, to expedite the initialization process, we offer warm-start options such as \texttt{CART} solutions in our algorithm. Besides, we set the mutation factor $M=rand(0,1)$ to enlarge the exploration range. Comprehensive \texttt{DEOCT} is detailed in Appendix \ref{apx:classicde}, Algorithm \ref{alg:de}.

\vskip -0.1in
\begin{minipage}[T]{\textwidth}
\setlength{\columnsep}{-5pt}
\begin{multicols}{2}
 \begin{minipage}[T]{0.48 \textwidth}
\begin{algorithm}[H]
    \small
    \begin{spacing}{1}
	\caption{GPU-parallelized OCT Fitness}
	\label{alg:oct_fit}
	\begin{algorithmic}[1]
		\STATE {\bfseries Input:} Training dataset $(\mathbf{X}, \mathbf{Y})$ with $n$ samples and $K$ classes, DE population $\mathcal{S}^g$ with $N$ individuals, number of branch nodes $SB$, number of leaf nodes $SL$, and minimum sample size $n_{min}$.
		\STATE {\bfseries Output:} Fitness value set $\{f(\mathbf{s}_r) \;|\; \mathbf{s}^g_r \in \mathcal{S}^g\}$.
		\STATE \textcolor{BrickRed}{\textbf{CPU}}: Decode trees $(\mathbf{a}, \mathbf{b})^g_r$ from $\mathbf{s}^g_r \in \mathcal{S}^g$ by \cref{eqn:aenc} and (\ref{eqn:benc}).
		\STATE \textcolor{BrickRed}{\textbf{H}}\textbf{2}\textcolor{mygreen}{\textbf{D}}: Transfer $(\mathbf{X}, \mathbf{Y})$ (if not exist on the device) and all the decoded trees to GPU global memory.
        \STATE \textcolor{mygreen}{\textbf{GPU}:} Initialize the number of samples per thread as $n_{stride}$; number of required threads as $\hat{n} \gets \lceil n/n_{stride} \rceil $; class flag matrix $\mathbf{Z} \gets \mathbf{0}^{\hat{n} \times (SL\times N) \times K}$.
        \STATE \textcolor{mygreen}{\textbf{begin GPU}} (\textcolor{mygreen}{\textbf{Kernel 1}} on GPU threads executed parallelly and asynchronously): 
        \STATE Read dataset $(\mathbf{X}^j, \mathbf{Y}^j)$ of thread $j$ with $n_{stride}$ adjacent samples from GPU global memory.  
        \FOR{$r$th tree $(\mathbf{a}, \mathbf{b})^g_r$ \textbf{in} all the decoded trees} 
        \FOR{$(\mathbf{x}^j_i, y_i^j)\in (\mathbf{X}^j, \mathbf{Y}^j)$}
        \STATE $t\gets 1$.
            \WHILE{$t \leq$ SB} 
            \STATE $t = x_{i, a_t}^j<b_t \; ? \; t\times2 \;:\; t\times2+1$.              
            \ENDWHILE
            \STATE $\mathbf{Z}[j,\; (r-1)\times SL + (t-SB),\; y_i^j] \; += \;1$.
        \ENDFOR
        \ENDFOR
        \STATE $\hat{\mathbf{Z}} =$ sum($\mathbf{Z}$, dims=1).
        \STATE \textcolor{mygreen}{\textbf{end GPU}}.
        \STATE \textcolor{BrickRed}{\textbf{D}}\textbf{2}\textcolor{mygreen}{\textbf{H}}: Transfer the class count matrix $\hat{\mathbf{Z}}$ to host
        \STATE \textcolor{BrickRed}{\textbf{CPU}}: Calculate the OCT cost of each individual $f(\mathbf{s}_r)$.
	\end{algorithmic}
    \end{spacing}
\end{algorithm}
\end{minipage}

\begin{minipage}[T]{0.5\textwidth}
\begin{algorithm}[H]
    \small
    \begin{spacing}{1.017}
	\caption{Moving-Horizon DEOCT} 
	\label{alg:ladeoct}
	\begin{algorithmic}[1]
		\STATE {\bfseries Input:} Training set $(\mathbf{X}, \mathbf{Y})$, tree depth $D$, minimum leaf sample number $n_{min}$, complexity parameter $\alpha$, moving-horizon depth $D_{\text{MH}}$. DE: warm-start trees $\{(\mathbf{a}, \mathbf{b})\}^0$, other parameters in DEOCT.
  	\STATE {\bfseries Output:} The best tree $(\mathbf{a}, \mathbf{b})_{best}$.
        \FOR{$t \in \mathcal{T}_B$}
            \STATE Induce the subset of training set for $t$th node $(\mathbf{X}_t, \mathbf{Y}_t)$ according to the path to node $t$.
            \IF{$|\mathbf{X}_t|>n_{min}$ and $\text{unique}(\mathbf{Y}_t)>1$}
                \STATE Determine the effective moving-horizon depth for $t$ node, $D_{t, \text{MH}} = \min\{D_{\text{MH}}, D-D_t\}$.
                    \IF{$D_{t, \text{MH}}\neq 1$}
                    \STATE Retrieve the warm-start subtrees $\{(\mathbf{a}, \mathbf{b})\}_t^0$ from the warm-start trees $\{(\mathbf{a}, \mathbf{b})\}^0$ with the same shape and location as the corresponding moving-horizon subtree.
                    \STATE Calculate the CART solution for the MH subtree and add to the warm-start subtrees.
                    \STATE Optimize MH subtree with depth $D_{t, \text{MH}}$ using DEOCT Alg.\ref{alg:de} and warm-start subtrees.
                    \ELSE
                    \STATE Optimize the MH subtree with depth $1$ using CART with the misclassification loss.
                    \ENDIF
                \STATE Save the first node of the optimized MH subtree in the $t$th node of $(\mathbf{a}, \mathbf{b})_{best}$.
            \ELSE
                \STATE Save an artificial node with $\{a_t=0, b_t=0\}$ in the $t$th node of $(\mathbf{a}, \mathbf{b})_{best}$.
            \ENDIF
        \ENDFOR		
	\end{algorithmic}
 \end{spacing}
\end{algorithm}
\end{minipage}
\end{multicols}
\end{minipage}
\vskip -0.05in

\section{Moving-Horizon DEOCT}
In a binary tree, the quantity of branch nodes grows exponentially with the tree depth $D$. \texttt{DEOCT}, which optimizes the entire tree simultaneously, leads to individual sizes that also expand exponentially at a rate of $O(2^{D+1}-2)$. This causes substantial computational costs for deep trees (e.g., depth=8), leading to suboptimal performance. Our tests indicate that \texttt{DEOCT}, if warm-started with \texttt{CART}, slightly outperforms standalone \texttt{CART} at a tree depth of 8. However, without a \texttt{CART} warm-start, \texttt{DEOCT} underperforms compared to \texttt{CART} at the same depth. 

\begin{figure}[!htp]
	\begin{center}
		\centerline{\includegraphics[width=0.8\columnwidth]{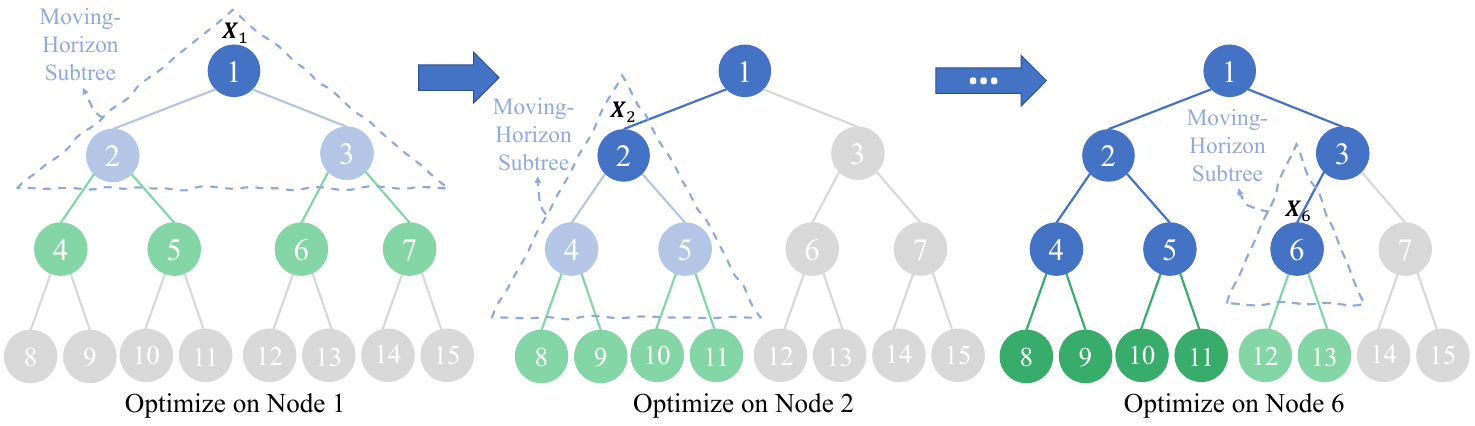}} 
		\caption{Training procedure of Moving-Horizon DEOCT (Depth=3, Moving-Horizon Depth=2)}
		\label{fig:LA}
	\end{center}
  \vskip -0.2in
\end{figure}

To tackle the challenge of rapidly growing node numbers and individual sizes with increasing tree depth, we propose a Moving-Horizon (MH) strategy. Similar to greedy approaches, MH optimizes node by node, but it optimizes a multi-layer subtree rooted at each node (called moving-horizon subtree) and adopts its first node in the final tree. While several moving-horizon algorithms for decision trees have been suggested in previous literature \citep{murthy1995lookahead, dong2001, esmeir2004, roizman2006, last2013}, they typically optimize the moving-horizon subtrees by employing greedy methods, such as \texttt{CART}, supplemented with additional post-processing. However, they generally achieve only marginal performance improvements over purely greedy methods. In contrast, our method, referred to as \texttt{Moving-Horizon DEOCT} (\texttt{MH-DEOCT}) in \cref{alg:ladeoct}, employs \texttt{DEOCT} to optimize the entire moving-horizon subtrees simultaneously. This approach provides a more comprehensive global view within a moving-horizon subtree, resulting in improved performance compared to both the purely greedy methods and the \texttt{DEOCT} methods. 

\cref{fig:LA} illustrates the training procedure for a tree with depth=3 and moving-horizon depth=2. The effective moving-horizon depth of the $t$th node ($D_{t, \text{MH}}$) is dynamically determined based on node depth $D_t$, tree depth $D$, and moving-horizon depth $D_{\text{MH}}$ using $D_{t, \text{MH}} = \min\{D_{\text{MH}}, D-D_t\}$. For example, when optimizing node 6 in \cref{fig:LA}, the effective moving-horizon depth is 1, as the child nodes' absolute depth in the subtree reaches the original tree's maximum depth.

\section{Numerical Results}
In this section, we evaluate the proposed algorithms, \texttt{DEOCT} and \texttt{MH-DEOCT}, alongside the classical heuristic method, \texttt{CART} \citep{breiman1984} and the state-of-the-art heuristic method, local search OCT (\texttt{LS-OCT}) \citep{bertsimasOptimal2017}, as well as state-of-the-art global optimal methods, namely \texttt{RS-OCT} \citep{hua2022a} and \texttt{Quant-BnB} \citep{mazumder2022} on continuous datasets and \texttt{GOSDT-Guess} \citep{mctavish2022a} on binary datasets. Our analysis addresses the following questions: \hyperref[Q1]{\textbf{Q1}}.How do our algorithms compare to heuristic and global optimal methods in terms of training accuracy? \hyperref[Q2]{\textbf{Q2}}.Can conclusions drawn from training accuracy be extended to testing accuracy? \hyperref[Q3]{\textbf{Q3}}.How does the training time of our algorithms compare to other methods with respect to dataset size and tree depth? \hyperref[Q4]{\textbf{Q4}}.What is the impact of moving-horizon depth on performance? \hyperref[Q5]{\textbf{Q5}}.How does the warm-start influence performance, and can our methods outperform heuristic methods without any warm-starts? \hyperref[Q6]{\textbf{Q6}}.How do our algorithms behave on binary datasets comparing to global methods designed for binary datasets?

\textbf{Experiments }We utilize 68 UCI classification datasets \citep{Dua:2019} with continuous and categorical variables for binary and multi-class tasks, and a sample size range of 47 to 11,000,000. Datasets are split into 75\% training and 25\% testing sets, repeated 10 times with varied random seeds. All the following results are complied from these repetitions. The computations are performed on a server with a 24-core AMD 2.65GHz CPU, 498 GB RAM, and an NVIDIA A100 (40 GB) GPU. For tree hyperparameters, \texttt{Quant-BnB} disregards minimum sample size ($n_{min}$) and complexity parameters ($\alpha$). To make the comparison fair, unless stated otherwise, we set $n_{min}$ to 1 and $\alpha$ to 0 for all algorithms. Additionally, we also provides the results of $\alpha$ tuning with validation sets in \cref{tbl:comparision} and \cref{apx:tuning}.

The proposed \texttt{DEOCT} and \texttt{MH-DEOCT} are implemented in \texttt{Julia} and run in parallel on a GPU using \texttt{CUDA.jl} \citep{besard2018juliagpu, cook2012}. The implementations will be open-sourced after acceptance. Notably, \texttt{DEOCT} utilize \texttt{CART} solutions as warm-starts;  \texttt{MH-DEOCT} utilize \texttt{CART} and \texttt{DEOCT} solutions as warm-starts in solving the MH subtrees as described in \cref{alg:ladeoct}. The recorded training time of our algorithms include the calculation time to obtain the warm-starts. As for the parameter for DE, we follow the general guidelines in \cite{ahmad_differential_2022}: crossover probability ($CR$) is set to 0.1, maximum generations ($G$) to 600, and population size ($N$) to 100 in Normal mode ($G$ increased to 4,000 and $N$ to 200 in Long mode; unless otherwise stated, all DE-related experiments run in Normal mode). We highlight that this parameter setting is used for all datasets. Comparison methods include \texttt{CART} using \texttt{DecisionTree.jl} \citep{ben_sadeghi_2022_7359268} running in the serial mode; \texttt{LS-OCT}, self-implemented in \texttt{Julia} in serial mode as no open-source code is provided; \texttt{Quant-BnB} from the official repository, executed in \texttt{Julia} in serial with a four-hour limit; and \texttt{RS-OCT} from the official repository in \texttt{Julia} with \texttt{CPLEX}, running on 40 CPU threads in parallel with a four-hour limit.

\subsection{Comparison with heuristic and global optimal methods} \label{sec:com_acc}
\vskip -0.05in
\begin{table}[!htp]
\caption{Comparison with heuristic and global methods on 65 datasets within 1 million samples}
\vskip -0.15in
\tabcolsep=5pt
\label{tbl:comparision}
\begin{center}
\renewcommand{\arraystretch}{0.9}
\resizebox{\textwidth}{!}{
\begin{threeparttable}
\begin{tabular}{c|c|cc|cc|cccc|c}
\toprule
\multirow{2}{*}{\begin{tabular}[c]{@{}c@{}}Perfor-\\ mance\end{tabular}} & \multirow{2}{*}{Depth} & \multicolumn{2}{c|}{Heuristic Method} & \multicolumn{2}{c|}{Global Method} & \multicolumn{5}{c}{Our Method} \\ \cline{3-11} 
 &  & CART\tnote{4} & LS-OCT\tnote{4} & Quant-BnB\tnote{4} & RS-OCT\tnote{4} & DEOCT\tnote{4} & MH-DEOCT\tnote{4} & \begin{tabular}[c]{@{}c@{}}DEOCT\tnote{4}\\      (Long)\end{tabular} & \begin{tabular}[c]{@{}c@{}}MH-DEOCT\tnote{4}\\      (Long)\end{tabular} & \begin{tabular}[c]{@{}c@{}}MH-DEOCT\\      ($\alpha$ tuned)\end{tabular} \\ \midrule
\multirow{4}{*}{\begin{tabular}[c]{@{}c@{}}Train\\      (\%)\end{tabular}} & 2 & 79.41 & 81.47 & \textbf{81.86}\tnote{1} & 81.42\tnote{3} & 81.56 & 81.64 & 81.66 & 81.75 & 81.25 \\
 & 3 & \begin{tabular}[c]{@{}c@{}}84.30\\      (84.05)\end{tabular} & \begin{tabular}[c]{@{}c@{}}86.04\\      (85.78)\end{tabular} & \begin{tabular}[c]{@{}c@{}}\textbf{87.37}\tnote{2}\\      \textbf{(87.43)}\end{tabular} & - & \begin{tabular}[c]{@{}c@{}}86.43\\      (86.19)\end{tabular} & \begin{tabular}[c]{@{}c@{}}86.82\\      (86.54)\end{tabular} & \begin{tabular}[c]{@{}c@{}}87.03\\      (86.75)\end{tabular} & \begin{tabular}[c]{@{}c@{}}87.23\\      (86.93)\end{tabular} & \begin{tabular}[c]{@{}c@{}}86.26\\      (85.92)\end{tabular} \\
 & 4 & 87.62 & 88.87 & - & - & 89.33 & 90.24 & 90.26 & \textbf{90.80} & 89.19 \\
 & 8 & 95.67 & 96.26 & - & - & 95.82 & 97.30 & 96.14 & \textbf{97.66} & 94.57 \\ \midrule
\multirow{4}{*}{\begin{tabular}[c]{@{}c@{}}Test\\      (\%)\end{tabular}} & 2 & 76.91 & 78.19 & 78.20 & 78.23 & 78.20 & 78.21 & 78.15 & \textbf{78.27} & \textbf{78.85} \\
 & 3 & \begin{tabular}[c]{@{}c@{}}80.45\\      (80.02)\end{tabular} & \begin{tabular}[c]{@{}c@{}}81.57\\      (81.10)\end{tabular} & \textbf{\begin{tabular}[c]{@{}c@{}}81.90\\      (81.72)\end{tabular}} & - & \begin{tabular}[c]{@{}c@{}}81.48\\      (81.09)\end{tabular} & \begin{tabular}[c]{@{}c@{}}81.78\\      (81.31)\end{tabular} & \begin{tabular}[c]{@{}c@{}}81.64\\      (81.25)\end{tabular} & \begin{tabular}[c]{@{}c@{}}81.74\\      (81.32)\end{tabular} & \textbf{\begin{tabular}[c]{@{}c@{}}82.71\\      (82.26)\end{tabular}} \\
 & 4 & 82.42 & 83.32 & - & - & 83.45 & 83.60 & 83.63 & \textbf{83.74} & \textbf{84.81} \\
 & 8 & 85.78 & 85.80 & - & - & 86.12 & \textbf{86.31} & 86.17 & 86.26 & \textbf{87.73} \\ \midrule
\multirow{4}{*}{\begin{tabular}[c]{@{}c@{}}Time\\      (s)\end{tabular}} & 2 & 0.03 & 140.34 & 2.96 & 7,701.63 & 0.72 & 1.51 & 8.54 & 17.17 & 29.46 \\
 & 3 & 0.05 & 231.98 & \begin{tabular}[c]{@{}c@{}}2204.39\\      (1059.91)\end{tabular} & - & 1.15 & 3.58 & 14.46 & 43.14 & 69.56 \\
 & 4 & 0.06 & 366.77 & - & - & 1.81 & 6.96 & 22.99 & 84.32 & 127.38 \\
 & 8 & 0.12 & 869.29 & - & - & 14.71 & 68.77 & 178.33 & 731.07 & 973.11 \\ \bottomrule
\end{tabular}
\begin{tablenotes}
    \item[1] All datasets achieve global optimality at Depth 2 in \texttt{Quant-BnB}. $^2$ 55/65 datasets reach global optimality at Depth 3 in \texttt{Quant-BnB}. $(\cdot)$ is the average accuracy or time of these 55 datasets. Besides, \texttt{Quant-BnB} failed to give a solution for 5/65 datasets.  $^3$ 35/65 datasets attain global optimality at Depth 2 in \texttt{RS-OCT}. $^4$ $\alpha$ is set to 0 without $\alpha$ tuning. 
\end{tablenotes}
\end{threeparttable}
}
\end{center}
\vskip -0.1in
\end{table}

\Cref{tbl:comparision} compares accuracy and training times for 65 datasets with up to 1 million samples. In terms of training accuracy (\label{Q1}\textbf{Q1}), \texttt{MH-DEOCT} outperforms heuristic methods \texttt{CART} and \texttt{LS-OCT} by average relative differences of 2.99\% and 1.05\% across all tree depths and datasets, respectively, and this performance gap extends to 3.44\% and 1.48\% in Long mode.  The global method \texttt{Quant-BnB} reaches global optima on 65 and 55 datasets at depths 2 and 3, respectively, while \texttt{MH-DEOCT} trails by 0.27\% and 1.09\%. Remarkably, given more time, \texttt{MH-DEOCT} (Long) achieves closer accuracy to \texttt{Quant-BnB} with only 0.14\% and 0.63\% differences, respectively. Besides, \texttt{Quant-BnB} failed to give a solution for 5 out of 65 datasets. For trees deeper than 3, \texttt{MH-DEOCT} continues to outperform heuristic methods  in a reasonable time, while \texttt{Quant-BnB} cannot solve these deeper trees. Meanwhile, \texttt{RS-OCT} reaches global optima on 35 datasets at depth 2. It lags behind \texttt{MH-DEOCT} in terms of average accuracy on all datasets and fails to perform at depths larger than 2.

\begin{figure}[!htp]
	\begin{center}
		\centerline{\includegraphics[width=0.95\columnwidth]{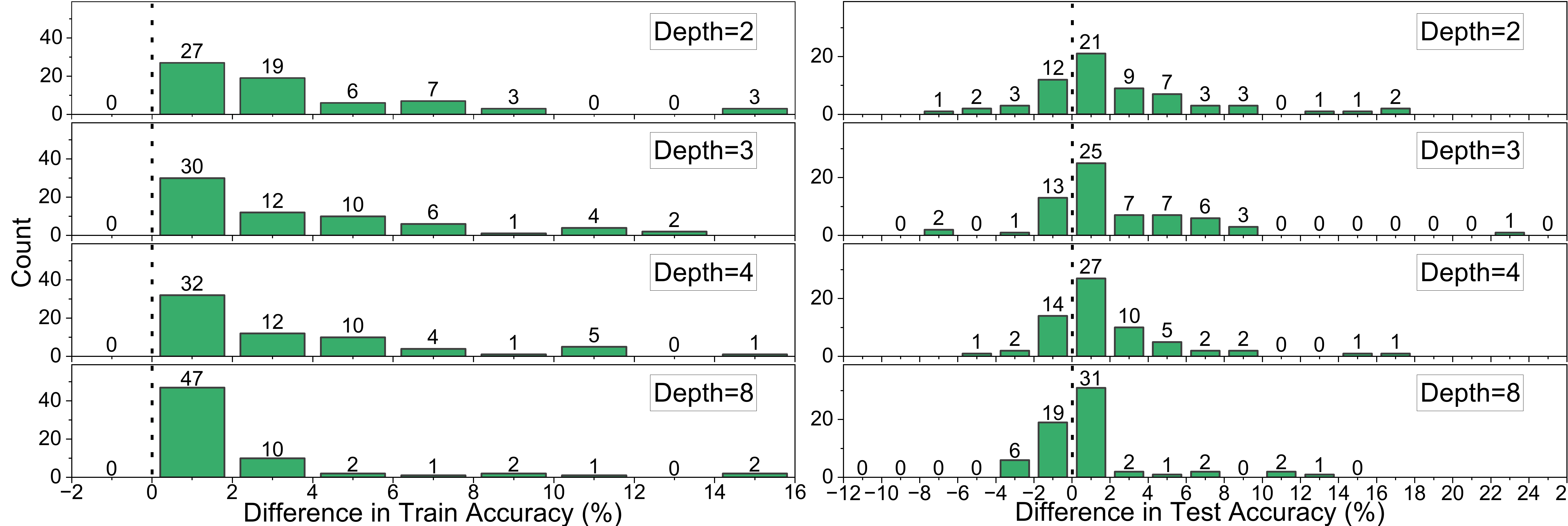}} 
		\caption{Comparison of \texttt{MH-DEOCT} with \texttt{CART} in relative difference ($\frac{\text{Acc}_{ _{\text{MH-DEOCT}}}-\text{Acc}_{ _\text{{CART}}}}{\text{Acc}_{ _\text{{CART}}}}\times 100 \%$) on 65 datasets within 1 million samples}
		\label{fig:delta}
	\end{center}
\end{figure}

Regarding testing accuracy (\label{Q2}\textbf{Q2}), \texttt{MH-DEOCT} exceeds \texttt{CART} and \texttt{LS-OCT} by average relative difference of 1.65\% and 0.35\% across all depths (1.71\% and 0.40\% in Long mode), respectively. \texttt{MH-DEOCT} marginally outperforms \texttt{Quant-BnB} at depth 2 and slightly underperforms at depth 3. \Cref{fig:delta} displays the number of datasets with relative differences between \texttt{MH-DEOCT} and \texttt{CART}. As depicted, \texttt{MH-DEOCT} outshines \texttt{CART} on most datasets for all depths. Detailed results on all datasets for \cref{tbl:comparision} are provided in \cref{apx:results}. 

\textbf{Moreover}, we present outcomes of hyperparameter $\alpha$ tuning using validation sets in \cref{tbl:comparision} and \cref{apx:tuning}. The parameter $\alpha$ plays a crucial role in balancing the tree node number (complexity) and the training accuracy. After tuning, \texttt{MH-DEOCT} improves its testing accuracy by an average of $1.37\%$ relatively, despite a 1.29\% drop in training accuracy. Tuned \texttt{MH-DEOCT} also outperforms tuned \texttt{CART} and \texttt{IAI-OCT} by 1.92\% and 0.80\%, respectively. Here, \texttt{IAI-OCT} is a commercial OCT method with hyperparameter $\alpha$ tuning \citep{bertsimasOptimal2017, InterpretableAI}.

\vskip -0.15in

\begin{figure}[!htp]
	\begin{center}
		\centerline{\includegraphics[width=0.95\columnwidth]{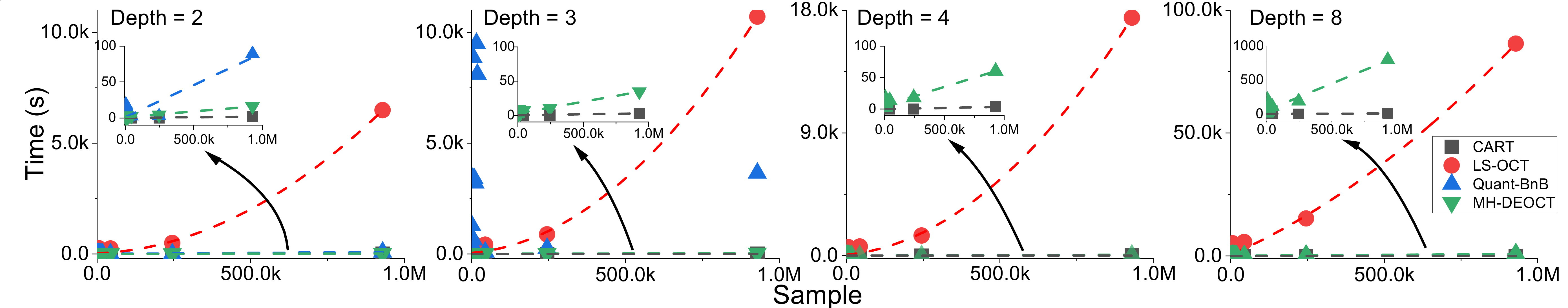}} 
         \vskip -0.05in
		\caption{Comparison of the training times regarding to the dataset sizes}
		\label{fig:time_results}
	\end{center}
  \vskip -0.25in
\end{figure}

Concerning training time (\label{Q3}\textbf{Q3}), \texttt{CART} performs the fastest, albeit at the expense of performance quality, while \texttt{MH-DEOCT} provides a balance between time efficiency and performance. Figure \ref{fig:time_results} illustrates the training time in relation to dataset sizes. Generally, \texttt{CART} and \texttt{MH-DEOCT} exhibit linear growth as sample numbers increase. \texttt{LS-OCT} shows an accelerated growth pattern. \texttt{Quant-BnB}, on the other hand, demonstrates a seemingly linear trend at depth 2, but becomes less stable at depth 3.

We also present comparisons with a global method requiring binary datasets, \texttt{GOSDT-Guess} \citep{mctavish2022a} (\label{Q6}\textbf{Q6}). As outlined in \cref{apx:binaryglobal}. Our algorithm, \texttt{MH-DEOCT}, performs commendably, with only a marginal decrease of 0.10\% and 0.12\% in average train and test accuracy when compared to the global methods on binarized datasets. Additionally, we observe that the global methods encounter challenges in obtaining optimal solutions within a reasonable time for large datasets and deepe trees. In contrast, our algorithm adeptly manages these complexities, efficiently solving problems involving larger datasets (up to 11,000,000 samples) and deeper trees, marking a groundbreaking achievement in the field. Comparing the outcomes of continuous and binarized datasets, our results highlight the superiority of continuous datasets in terms of average train and test accuracy when contrasted with their binarized counterparts. This disparity suggests a potential information loss during the binarization process, aligning with prior observations in \citep{mazumder2022}. Consequently, \texttt{MH-DEOCT} consistently attains higher average accuracy on continuous datasets in comparison to the global methods applied to their corresponding binarized equivalents.

\subsection{Ablation experiments} \label{sec:ablation}
\begin{figure}[!htp]
	\begin{center}
		\centerline{\includegraphics[width=0.85\columnwidth]{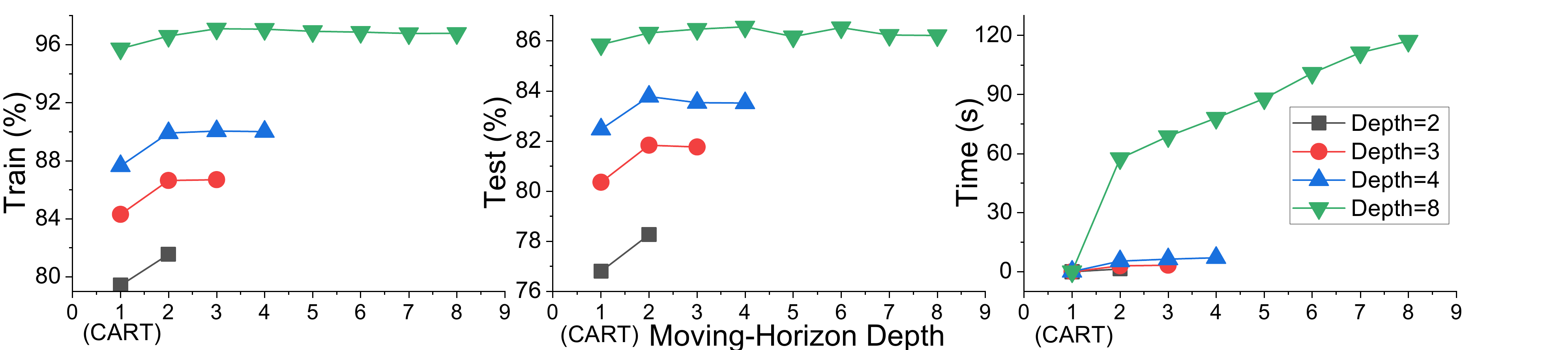}} 
		\caption{Influence of Moving-Horizon Depth on accuracy and training times}
		\label{fig:scalability}
	\end{center}
  \vskip -0.1in
\end{figure}
  \vskip -0.1in
\cref{fig:scalability} illustrates the effect of moving-horizon (MH) depth on accuracy and training times in the normal mode (\label{Q4}\textbf{Q4}). Training accuracy generally improves for MH depths up to 3 and stabilizes or slightly decreases beyond. Training time increases with MH depth, thus we choose MH depth 2 for tree depth 2 and MH depth 3 for deeper trees as the optimal MH depth. This observation may be due to the fact that, under the normal mode, \texttt{DEOCT} can only yield reasonable solutions for depths $\leq 3$. Larger generation and population sizes could yield a larger optimal MH depth. Moreover, by comparing the training and testing accuracy plots, we observe that testing accuracy increases proportionally to training accuracy at depths 2, 3, and 4, and to a lesser extent at depth 8. This result suggests that our algorithms do not suffer extensively from overfitting in deeper trees.

\begin{table}[!htp]
\caption{Ablation study on DEOCT and MH-DEOCT with/without warm-starts (WS)}
\label{tbl:ablation}
\begin{center}
\renewcommand{\arraystretch}{0.95}
\begin{threeparttable}
\resizebox{0.85\textwidth}{!}{
\begin{tabular}{c|ccc|ccc|ccc|ccc}
\toprule
\multirow{2}{*}{No.} & \multicolumn{3}{c|}{\multirow{2}{*}{Methods   \& Factors}} & \multicolumn{3}{c|}{D=2} & \multicolumn{3}{c|}{D=4} & \multicolumn{3}{c}{D=8} \\
 & \multicolumn{3}{c|}{} & \begin{tabular}[c|]{@{}c@{}}Train\\      (\%)\end{tabular} & \begin{tabular}[c|]{@{}c@{}}Test\\      (\%)\end{tabular} & \begin{tabular}[c]{@{}c@{}}Time\\       (s)\end{tabular} & \begin{tabular}[c]{@{}c@{}}Train\\      (\%)\end{tabular} & \begin{tabular}[c]{@{}c@{}}Test\\      (\%)\end{tabular} & \begin{tabular}[c]{@{}c@{}}Time\\       (s)\end{tabular} & \begin{tabular}[c]{@{}c@{}}Train\\      (\%)\end{tabular} & \begin{tabular}[c]{@{}c@{}}Test\\      (\%)\end{tabular} & \begin{tabular}[c]{@{}c@{}}Time\\       (s)\end{tabular} \\
\midrule
1 & \multicolumn{3}{c|}{\textbf{CART}} & 79.41 & 76.91 & 0.03 & 87.62 & 82.42 & 0.06 & 95.67 & 85.78 & 0.12 \\
\midrule
 & \multicolumn{3}{c|}{\textbf{DEOCT}} & \multicolumn{3}{c|}{\multirow{2}{*}{D=2}} & \multicolumn{3}{c|}{\multirow{2}{*}{D=4}} & \multicolumn{3}{c}{\multirow{2}{*}{D=8}} \\
 & \multicolumn{3}{c|}{CART WS in DE} & \multicolumn{3}{c|}{} & \multicolumn{3}{c|}{} & \multicolumn{3}{c}{} \\
\cline{1-13} 
2 & \multicolumn{3}{c|}{-} & 81.50 & 78.31 & 0.70 & 88.45 & 83.27 & 1.63 & 91.22 & 84.51 & 14.91 \\
3 & \multicolumn{3}{c|}{$\checkmark$} & 81.56 & 78.20 & 0.72 & 89.33 & 83.45 & 1.81 & 95.82 & 86.12 & 14.71 \\
4 & \multicolumn{3}{c|}{- (extended)} & - & - & - & 89.26 & 83.60 & 4.87 & 92.33 & 85.25 & 44.17 \\
5 & \multicolumn{3}{c|}{$\checkmark$ (extended)} & - & - & - & 89.49 & 83.73 & 5.08 & 96.02 & 86.36 & 39.74 \\
\midrule
 & \multicolumn{3}{c|}{\textbf{MH-DEOCT}} & \multicolumn{3}{c|}{D=2, P=2} & \multicolumn{3}{c|}{D=4, P=3} & \multicolumn{3}{c}{D=8, P=3} \\
 & \begin{tabular}[c]{@{}c@{}}CART\\      WS in DE\end{tabular} & \begin{tabular}[c]{@{}c@{}}DE\\      WS\end{tabular} & \begin{tabular}[c]{@{}c@{}}CART\\      WS\end{tabular} & \begin{tabular}[c]{@{}c@{}}Train\\      (\%)\end{tabular} & \begin{tabular}[c]{@{}c@{}}Test\\      (\%)\end{tabular} & \begin{tabular}[c]{@{}c@{}}Time\\       (s)\end{tabular} & \begin{tabular}[c]{@{}c@{}}Train\\      (\%)\end{tabular} & \begin{tabular}[c]{@{}c@{}}Test\\      (\%)\end{tabular} & \begin{tabular}[c]{@{}c@{}}Time\\       (s)\end{tabular} & \begin{tabular}[c]{@{}c@{}}Train\\      (\%)\end{tabular} & \begin{tabular}[c]{@{}c@{}}Test\\      (\%)\end{tabular} & \begin{tabular}[c]{@{}c@{}}Time\\       (s)\end{tabular} \\
 \cline{1-13}
6 & - & - & - & 81.51 & 78.14 & 0.71 & 89.85 & 83.57 & 4.69 & 96.72 & 86.36 & 33.75 \\
7 & - & - & $\checkmark$ & 81.40 & 78.23 & 0.76 & 89.93 & 83.57 & 4.93 & 96.84 & 86.42 & 32.68 \\
8 & - & $\checkmark$ & - & 81.61 & 78.18 & 1.40 & 90.01 & 83.58 & 6.35 & 96.71 & 86.42 & 57.97 \\
9 & - & $\checkmark$ & $\checkmark$ & 81.63 & 78.29 & 1.50 & 90.03 & 83.71 & 6.39 & 96.93 & \textbf{86.51} & 55.91 \\
10 & $\checkmark$ & $\checkmark$ & - & 81.57 & \textbf{78.32} & 1.45 & 90.09 & \textbf{83.74} & 6.48 & 97.04 & 86.30 & 57.48 \\
11 & $\checkmark$ & $\checkmark$ & $\checkmark$ & \textbf{81.64} & 78.21 & 1.51 & \textbf{90.24} & 83.60 & 6.96 & \textbf{97.30} & 86.31 & 68.77 \\
\bottomrule
\end{tabular}
}
\end{threeparttable}
\end{center}
\vskip -0.15in
\end{table}

\cref{tbl:ablation} displays the impact of warm-starts (WS) on accuracy (\label{Q5}\textbf{Q5}). Several key observations can be drawn: first, from rows 1, 3, and 7, with \texttt{CART} warm-starts, both \texttt{DEOCT} and \texttt{MH-DEOCT} exhibit improved training accuracy compared to \texttt{CART}. Second, from rows 1, 2, and 6, even without any warm-start, \texttt{MH-DEOCT} consistently outperforms \texttt{CART} across all depths, while \texttt{DEOCT} falls behind at depth 8. Third, we extend the iterations of \texttt{DEOCT} to $1800$ to match the training time of \texttt{MH-DEOCT} without \texttt{DEOCT} warm-starts. Rows 4, 5, 6, and 7 demonstrates that \texttt{MH-DEOCT} without \texttt{DEOCT} warm-starts surpasses \texttt{DEOCT} while maintaining similar running times. Finally, from rows 8, 9, 10, and 11, \texttt{MH-DEOCT} with \texttt{DEOCT} warm-starts can achieves substantially better training accuracy compared to \texttt{DEOCT}, with the significance increasing as tree depth grows.

\subsection{Performance on large-scale Datasets (millions of samples)} \label{sec:huge}
\cref{tbl:comparision} presents the performance of \texttt{MH-DEOCT} on three large-scale datasets with millions of samples while all the global optimal methods and \texttt{LS-OCT} failed to give a result with reasonable running time. In general, \texttt{MH-DEOCT} demonstrates improved performance over \texttt{CART} in terms of both training and testing accuracy, with a trade-off in training time.

\begin{table}[!htp]
\caption{Performance of MH-DEOCT on large-scale datasets (millions of samples)}
\label{tbl:large}
\begin{center}
\renewcommand{\arraystretch}{0.9}
\begin{threeparttable}
\resizebox{0.85\textwidth}{!}{
\begin{tabular}{ccccccccccc}
\toprule
\multirow{2}{*}{Dataset} & \multirow{2}{*}{Sample} & \multirow{2}{*}{Feature} & \multirow{2}{*}{Class} & \multirow{2}{*}{\begin{tabular}[c]{@{}c@{}}Perfor-\\      mance\end{tabular}} & \multicolumn{2}{c}{D=2} & \multicolumn{2}{c}{D=4} & \multicolumn{2}{c}{D=8} \\
 &  &  &  &  & CART & \multicolumn{1}{l}{MH-DEOCT} & CART & \multicolumn{1}{l}{MH-DEOCT} & CART & MH-DEOCT \\
 \midrule
\multirow{3}{*}{\begin{tabular}[c]{@{}c@{}}Bitcoin\\      Heist\end{tabular}} & \multirow{3}{*}{2,916,697} & \multirow{3}{*}{8} & \multirow{3}{*}{29} & Train (\%) & 98.58 & 98.58 & 98.58 & 98.58 & 98.61 & \textbf{98.62} \\
 &  &  &  & Test (\%) & 98.58 & 98.58 & 98.58 & 98.58 & 98.61 & \textbf{98.62} \\
 &  &  &  & Time (s) & 3.64 & 77.22 & 5.91 & 259.98 & 9.59 & 1,633.93 \\
 \midrule
\multirow{3}{*}{SUSY} & \multirow{3}{*}{5,000,000} & \multirow{3}{*}{18} & \multirow{3}{*}{2} & Train (\%) & 74.46 & \textbf{75.85} & 76.59 & \textbf{77.79} & 78.38 & \textbf{79.32} \\
 &  &  &  & Test (\%) & 74.47 & \textbf{75.84} & 76.59 & \textbf{77.78} & 78.34 & \textbf{79.17} \\
 &  &  &  & Time (s) & 18.93 & 164.03 & 37.32 & 399.88 & 75.90 & 5,100.47 \\
 \midrule
\multirow{3}{*}{HIGGS} & \multirow{3}{*}{11,000,000} & \multirow{3}{*}{28} & \multirow{3}{*}{2} & Train (\%) & 63.08 & \textbf{63.49} & 65.58 & \textbf{66.71} & 69.47 & \textbf{70.50} \\
 &  &  &  & Test (\%) & 63.06 & \textbf{63.48} & 65.55 & \textbf{66.68} & 69.40 & \textbf{70.38} \\
 &  &  &  & Time (s) & 44.52 & 333.01 & 82.28 & 1,049.87 & 171.65 & 15,885.24 \\
 \bottomrule
\end{tabular}
}
\end{threeparttable}
\end{center}
\vskip -0.2in
\end{table}

\subsection{Speedup Analysis: Comparing the GPU and CPU Versions of MH-DEOCT}

\begin{figure}[!htp]
	\begin{center}
		\centerline{\includegraphics[width=0.75\columnwidth]{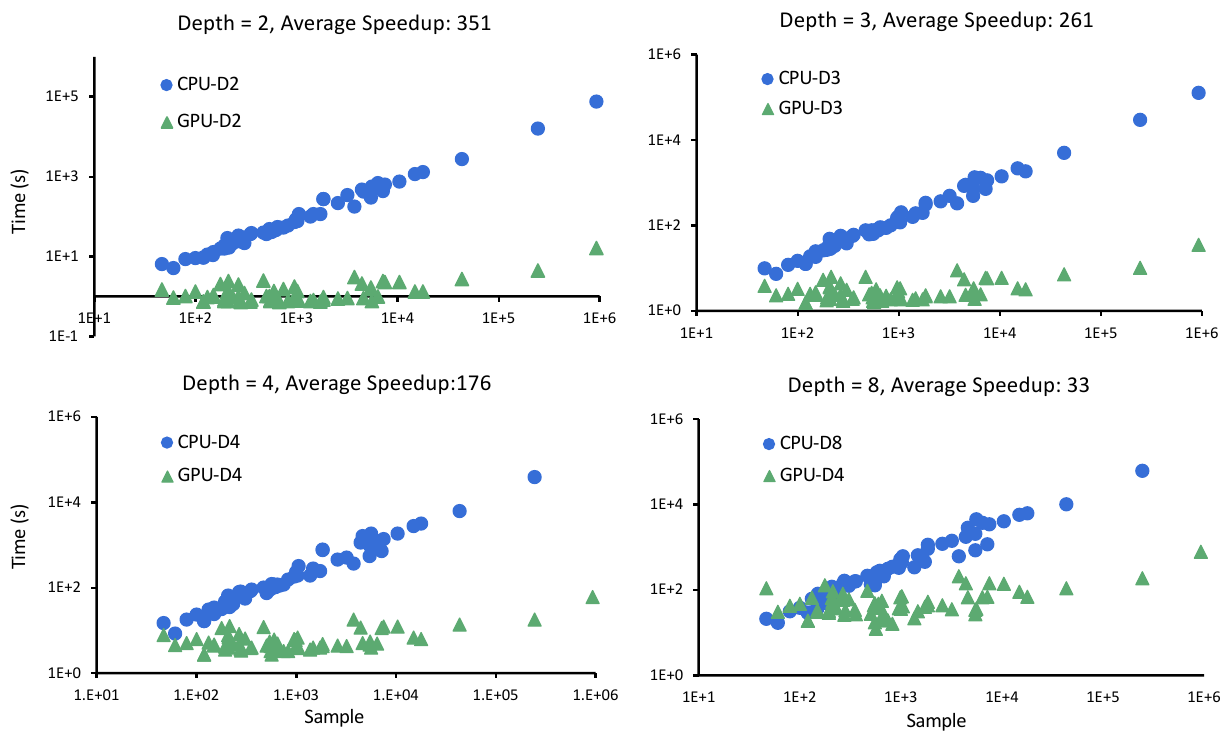}} 
		\caption{Speedup of MH-DEOCT on GPU and CPU over dataset size in log scale}
		\label{fig:speedup}
	\end{center}
  \vskip -0.1in
\end{figure}

To facilitate a comprehensive performance evaluation, we conducted an exhaustive assessment of the \texttt{MH-DEOCT} algorithm on both GPU and CPU platforms across 65 diverse datasets, ranging in size from up to 1 million samples. \cref{tbl:speedup} summarized results of these evaluations, presenting a quantified speedup factor of 205 on average across all tree depths and datasets. The performance disparity becomes particularly evident with different dataset sizes. For instance, datasets with sample sizes falling within the range of 0-10,000 exhibit a GPU speedup of 46 times compared to the CPU. As the dataset size escalates to the range of 10,000-250,000 samples, the speedup surges to an impressive 805 times. Most strikingly, when confronted with the extensive Htsensor dataset containing 928,991 samples, CPU couldn't produce results within a two-day limit for tree depths of 4 and 8, while the GPU excels by completing the calculations within 60 and 800 seconds, respectively. Additionally, a visual representation of the GPU's performance advantage over dataset size is provided in \cref{fig:speedup}.

\begin{table}[!htp]
\caption{Speedup of MH-DEOCT on GPU and CPU}
\label{tbl:speedup}
\renewcommand{\arraystretch}{0.9}
\center
\resizebox{0.7\textwidth}{!}{
\begin{tabular}{c|cccc|c}
\toprule
Dataset Size & \multicolumn{1}{l}{Depth=2} & \multicolumn{1}{l}{Depth=3} & \multicolumn{1}{l}{Depth=4} & \multicolumn{1}{l|}{Depth=8} & \multicolumn{1}{l}{Average on Size} \\ 
\midrule
0-10,000 & 79 & 56 & 38 & 9 & 46 \\
10,000-250,000 & 1359 & 1014 & 727 & 121 & 805 \\
\multicolumn{1}{c|}{928,991} & 4761 & 3671 & \textgreater{}2866 & \textgreater{}215 & - \\ \hline
Average    on Depth & 351 & 261 & 176 & 33 & 205 \\
\bottomrule
\end{tabular}
}
\end{table}

  \vskip -0.1in

\section{Conclusion}
In conclusion, we have proposed \texttt{MH-DEOCT}, a GPU-accelerated moving-horizon differential evolution algorithm for classification trees with continuous features. This approach addresses the limitations of existing methods by incorporating a discrete tree deocding method, a GPU-accelerated implementation, and a moving-horizon strategy for iteratively training shallow subtrees at each node. Our comprehensive evaluation on 68 UCI datasets demonstrates that \texttt{MH-DEOCT} outperforms heuristic methods and achieves a comparable accuracy level to global methods with significantly reduced running time. This performance can be extended to deeper trees and large-scale datasets. This work contributes to the ongoing effort to enhance decision tree training methodologies, offering a promising solution that combines the strengths of both heuristic and global approaches. 

Limitation: while our algorithm lacks a theoretical guarantee to reach global optima, its practical applications have demonstrated near-optimal performance, as evidenced by our extensive experiments.

\clearpage
\bibliography{deoct}
\bibliographystyle{unsrtnat}

\clearpage

\appendix
\section*{Appendix}
\input{appendix}
\end{document}

%% file: appendix.tex
\section{Overview of related works} \label{apx:related}
Over the years, \textbf{heuristic} techniques have dominated the domain of decision tree learning due to their inherent efficacy and scalability. Algorithms such as CART, proposed by Breiman et al. in 1984 [1], and C4.5, by Quinlan in 1993 [2], are prime examples. These methods initialize with a single node and iteratively expand the tree, often employing post-processing measures to prune the decision trees and minimize overfitting. These classic heuristic methods are deterministic, and once the procedure is completed, they are unable to generate a better solution. To overcome this limitation, a local search method introduced by Bertsimas and Dunn (2017) [3] aims to re-optimize each node by finding best splits at that specific node, while maintaining this node's original subtrees. It also introduces stochasticity by applying this procedure to multiple initial trees. Demonstrating superior performance compared to CART, this local search method is utilized as a baseline in our research. However, despite their widespread application, these heuristic techniques suffer from a significant drawback: they tend to concentrate solely on individual tree splits, overlooking potential consequences of subsequent splits [3]. As a result, these algorithms risk settling for sub-optimal solutions, which could impair the predictive effectiveness of the decision tree.
    
In response to the constraints posed by heuristic approaches when dealing with greedy problems, there has been a growing surge in the popularity of the optimal decision tree method. This approach aims to comprehensively explore the entire tree concurrently. Three primary categories of methods have gained prominence: mixed integer programming, dynamic programming, and metaheuristic techniques.
    
\textbf{Mixed-integer programming (MIP)} methods have recently emerged as powerful tools for constructing optimal decision trees. The fundamental characteristic of these methods is their ability to encode the entire tree by pre-determining the tree depth, establishing variables for each node predicate, and instituting constraints to maintain the tree's structural integrity. On the frontier of applying MIP to decision tree construction, Bertsimas and Shioda's work [4] in 2007 stood out, particularly suited for smaller datasets. Their pioneering work paved the way for enhanced MIP formulations by Bertsimas and Dunn (2017) [3] and Verwer and Zhang (2017) [5]. Verwer and Zhang took a significant leap forward in 2019 with the introduction of BinOPT, which optimized the MIP approach by using binary data [6]. This innovation reduced the number of necessary variables and constraints. 
    
While MIP is flexible and capable of integrating various constraints and losses, its inherent complexity often makes it difficult to solve. Among the state-of-the-art solutions addressing the complexity problems, RS-OCT by Hua et al. (2022) [7] has transformed the MIP formulation into a two-stage format with tree structure variables designated as the first-stage variables. RS-OCT argues that branching solely on these first-stage variables is sufficient to ensure convergence. Additionally, they introduced a closed-form lower bound along with two lower bounds based on general-purpose solvers. Meanwhile, Quant-BnB by Mazumder et al. (2022) [8] employs a quantile-based branching and lower bounding strategy, complemented by an efficient single-layer exhaustive search solver. Although these methods offer a theoretical guarantee of reaching the global optimum, they still incur significant computational expenses and are typically limited to shallow decision trees with depths of 2 or 3.
    
\textbf{Dynamic programming (DP)}, a method that breaks problems down into smaller, simpler subproblems, offers a robust solution to constructing optimal decision trees. When applied to decision trees, DP operates on two crucial observations. First, the left and right subtrees stemming from any given node can be optimized separately. This characteristic allows complex tree problems to be fragmented into more manageable subproblems. The second observation highlights overlaps within the tree. This overlap means certain subtrees can be reused during the solution process, making exhaustive searches more efficient through caching techniques. This methodological framework, termed DL8, was initially put forth by Nijssen and Fromont (2007, 2010) [9, 10]. This was further built upon by Aglin et al. (2020) [11] with DL8.5, which combined previous methodologies to optimize misclassification scores while considering tree depth. Building upon the DL8 framework, DL8.5 brought forth innovative bounding techniques. Specifically, the upper bounding method in DL8.5 recalibrated the upper bound of a child node using its parent's upper bound combined with the optimal value of its sibling. In tandem, DL8.5 also introduced a new lower bound, derived directly from the associated upper bound. Pursuing this trajectory, Demirović et al. (2022) [12] presented MurTree, which aimed to refine misclassification scores while considering constraints on tree depth and the number of nodes. MurTree came with a slew of enhancements, including a frequency-based solver for depth-two trees, incremental lower bounds, and a caching system tailored for optimal solutions and lower bounds using sub-datasets. These cumulative improvements endowed MurTree with superior scalability compared to its predecessors.

In addition to setting strict limitations on tree depth and nodes, Hu et al. (2019) [13] unveiled an algorithm that harmonizes between misclassifications and tree complexity by incorporating a regularization term into the cost function. Their approach leans heavily on exhaustive search, caching, and establishing a lower bound for misclassifications that's determined by the cost of adding a fresh node to the decision tree. Building on this, Lin et al. (2020) [14] fine-tuned and expanded upon the method, achieving commendable results when the complexity coefficient is set notably high. Based on Lin's work, McTavish et al. (2022) [15] utilized a pre-structured ensemble model to give a more informed estimation regarding binarization techniques, tree dimensions, and lower bounds. Leveraging these insights, their refined algorithm demonstrated marked enhancements in speed and efficiency. However, it's worth noting that these dynamic programming strategies are best suited for binary datasets. Challenges crop up when tackling continuous datasets, such as potential information degradation during binarization, an augmented feature space post-binarization, and escalating computational demands correlating with feature quantity and tree depth.
        
\textbf{Meta-Heuristic} approaches aim to identify the optimal solution by analyzing the entire tree. While these methods lack a theoretical guarantee of achieving global optimality, numerous applications have demonstrated their proficiency in finding satisfactory solutions. The inception of Meta-Heuristic based decision tree algorithms dates back to the early 1990s [16], and since then, substantial research has been undertaken. Researchers have primarily focused on enhancing performance by addressing two pivotal issues: encoding the decision tree into an individual vector and preserving the tree structure during mutation operations. For the encoding aspect, the prevalent approach is the linear representation, which captures the split feature and thresholds of each tree node in a sequential manner within the individual vector. References [17, 18] detail the ordered sequence of split features in the individual vector while dynamically searching for the corresponding split values. Conversely, [19] focuses on cataloguing the sequence of split thresholds, and [20] encode both the split feature and thresholds concurrently. In terms of mutation operations, there's a trend shifting from mutations based on absolute positions within individual vectors to those determined by relative positions in a decision tree. This includes strategies like node swaps at identical relative positions [21], subtree swaps at the same relative positions [22], and branch swaps sharing the same root node [23].

The foundational metaheuristic approach DEOCT in \textbf{our algorithm} is inspired by Meta-Heuristic methods that employ a linear representation for both split features and thresholds. It also uses a direct mutation based on absolute positions, which is a considerably simplified version compared to previous studies. Nonetheless, we've addressed two significant limitations inherent to Meta-Heuristic methods. Firstly, in the context of deeper trees, the performance of Meta-Heuristic methods typically falls short compared to the CART approach when not initialized with a CART warm-start, as confirmed by [24] and corroborated by our ablation study presented in Table 2 of the original paper. Secondly, Meta-Heuristic methods generally exhibit longer run-times relative to greedy techniques because they necessitate evaluating a vast set of potential solutions [20]. To mitigate the first issue, we implemented a moving horizon strategy that focuses on solving a shallow subtree problem. This strategy not only help keep the tree structure but also leverages the strengths of metaheuristics in handling shallow trees. For the second challenge, we introduced a GPU-accelerated version of the decision tree cost evaluation algorithm, complemented by a population-based differential evolution method optimized for GPU parallelism.

\subsection{References}

[1] Breiman, L., et al. "Classification and Regression Trees." (1984).

[2] Quinlan, J. Ross. "C 4.5: Programs for machine learning." The Morgan Kaufmann Series in Machine Learning (1993).

[3] Bertsimas, Dimitris, and Jack Dunn. "Optimal classification trees." Machine Learning 106 (2017): 1039-1082.

[4] Bertsimas, Dimitris, and Romy Shioda. "Classification and regression via integer optimization." Operations research 55.2 (2007): 252-271.

[5] Verwer, Sicco, and Yingqian Zhang. "Learning decision trees with flexible constraints and objectives using integer optimization." In Proceedings of CPAIOR (2017).

[6] Verwer, Sicco, and Yingqian Zhang. "Learning optimal classification trees using a binary linear program formulation." In Proceedings of AAAI (2019).

[7] Hua, Kaixun, Jiayang Ren, and Yankai Cao. "A Scalable Deterministic Global Optimization Algorithm for Training Optimal Decision Tree." In Proceedings of NeurIPS (2022).

[8] Mazumder, Rahul, Xiang Meng, and Haoyue Wang. "Quant-BnB: A Scalable Branch-and-Bound Method for Optimal Decision Trees with Continuous Features." In Proceedings of ICML (2022).

[9] Nijssen, Siegfried, and Elisa Fromont. "Mining optimal decision trees from itemset lattices." In Proceedings of SIGKDD (2007).

[10] Nijssen, Siegfried, and Elisa Fromont. "Optimal constraint-based decision tree induction from itemset lattices." Data Mining and Knowledge Discovery 21 (2010): 9-51.

[11] Aglin, Gaël, Siegfried Nijssen, and Pierre Schaus. "Learning optimal decision trees using caching branch-and-bound search." In Proceedings of AAAI (2020).

[12] Demirović, Emir, et al. "Murtree: Optimal decision trees via dynamic programming and search." The Journal of Machine Learning Research 23.1 (2022): 1169-1215.

[13] Hu, Xiyang, Cynthia Rudin, and Margo Seltzer. "Optimal sparse decision trees." In Proceedings of NeurIPS (2019).

[14] Lin, Jimmy, et al. "Generalized and scalable optimal sparse decision trees." In Proceedinhs of ICML (2020).

[15] McTavish, Hayden, et al. "Fast sparse decision tree optimization via reference ensembles." In Proceedings of AAAI (2022).

[16] Sutton, C. "Improving classification trees with simulated annealing." Computing Science and Statistics: Proceedings of the 23rd Symposium on the Interface. 1991.

[17] Cha, Sung-Hyuk, and Charles C. Tappert. "Constructing Binary Decision Trees using Genetic Algorithms." GEM. 2008.

[18] Cha, Sung-Hyuk, and Charles C. Tappert. "A genetic algorithm for constructing compact binary decision trees." Journal of pattern recognition research 4.1 (2009): 1-13.

[19] Omielan, Adam, and Sunil Vadera. "ECCO: A new evolutionary classifier with cost optimisation." Intelligent Information Processing VI: 7th IFIP TC 12 International Conference (2012).

[20] Rivera-Lopez, Rafael, and Juana Canul-Reich. "Construction of near-optimal axis-parallel decision trees using a differential-evolution-based approach." IEEE Access 6 (2018): 5548-5563.

[21] Sörensen, Kenneth, and Gerrit K. Janssens. "Data mining with genetic algorithms on binary trees." European Journal of Operational Research 151.2 (2003): 253-264.

[22] Papagelis, Athanassios, and Dimitrios Kalles. "GA Tree: genetically evolved decision trees." In Proceedings 12th IEEE ICTAI (2000).

[23] Fu, Zhiwei, et al. "Genetically engineered decision trees: population diversity produces smarter trees." Operations Research 51.6 (2003): 894-907.

[24] Ersoy, Elif, Erinç Albey, and Enis Kayis. "A CART-based Genetic Algorithm for Constructing Higher Accuracy Decision Trees." DATA. 2020.

\newpage
\section{Differential Evolution and DEOCT algorithms} \label{apx:classicde}
\begin{algorithm}[!htb]
	\caption{Classic Differential Evolution} 
	\label{alg:classicde}
	\begin{algorithmic}[1]
		\STATE {\bfseries Input:} The population size $N$, maximum generation $G$, mutation factor $M$, and cross probability $CR$.
        \STATE {\bfseries Output:}  The best individual $\mathbf{s}_{best}^g$.
		\STATE Initialize $g \gets 1, \mathcal{S}^g \gets \emptyset$.
		\FOR{$r \in {1,\cdots,N}$}
		\STATE Randomly initialize the individual $\mathbf{s}_r^g$ within the range limits for each variable.
		\STATE $\mathcal{S}^g \gets \mathcal{S}^g \cup \mathbf{s}_r^g$.
		\ENDFOR
		\STATE Update the best individual, $\mathbf{s}_{best}^g \gets \arg \min_{\mathbf{s}}f(\mathbf{s}_r^g),\; \mathbf{s}_r^g \in \mathcal{S}^g$.
		\WHILE{$g \leq$ maximum generation}
		\STATE $\mathcal{S}^{g+1} \gets \emptyset$.
		\FOR{$r\in{1,\cdots,N}$}
		\STATE Randomly select two individuals $\mathbf{s}_{r1}^{g}, \mathbf{s}_{r2}^{g}$ from $\mathcal{S}^{g}$.
		\STATE Generate the mutant vector $\mathbf{v}_r^{g}$ using \cref{eqn:mutation}.
		\STATE Generate the cross vector $\mathbf{u}_r^{g}$ using \cref{eqn:crossover}.
        \STATE Select the better vector $\mathbf{s}_r^{g+1}$ using \cref{eqn:selection}.
		\STATE $\mathcal{S}^{g+1} \gets \mathcal{S}^{g+1} \cup \mathbf{s}_r^{g+1}$.
		\ENDFOR
		\STATE Update the best individual, $\mathbf{s}_{best}^{g+1} \gets \arg \min_{\mathbf{s}} \{ \mathbf{s}_{best}^{g}, f(\mathbf{s}_r^{g+1})\},\; \mathbf{s}_r^{g+1}
		\in \mathcal{S}^{g+1}$.
        \STATE $g \gets g+1$.
		\ENDWHILE
	\end{algorithmic}
\end{algorithm}

\begin{algorithm}[H]
    \small
    \begin{spacing}{1}
	\caption{DEOCT} 
	\label{alg:de}
	\begin{algorithmic}[1]
		\STATE {\bfseries Input:} OCT: training set $(\mathbf{X}, \mathbf{Y})$, tree depth $D$, minimum sample size $n_{min}$, and complexity parameter $\alpha$. DE: the population size $N$, maximum generation $G$, warm-starts $\mathcal{S}^0$, mutation factor $M$, and cross probability $CR$.
  	\STATE {\bfseries Output:}  The best individual $\mathbf{s}_{best}^g$.
		\STATE Initialize $g \gets 1, \mathcal{S}^g \gets \mathcal{S}^0$.
		\FOR{$r \in {|\mathcal{S}^0|+1,\cdots,N}$}
		\STATE Randomly initialize the individual $\mathbf{s}_r^g$ within the range limits for each variable.
		\STATE $\mathcal{S}^g \gets \mathcal{S}^g \cup \mathbf{s}_r^g$.
		\ENDFOR
		\STATE $\mathbf{s}_{best}^g \gets \arg \min_{\mathbf{s}}f(\mathbf{s}_r^g),\; \mathbf{s}_r^g \in \mathcal{S}^g$.
		\WHILE{$g \leq G$}
		\STATE $\mathcal{S}^{g+1} \gets \emptyset$, $\mathcal{U}^g \gets \emptyset$.
		\FOR{$r\in{1,\cdots,N}$}
		\STATE Randomly select $\mathbf{s}_{r1}^{g}, \mathbf{s}_{r2}^{g}$ from $\mathcal{S}^{g}$.
		\STATE Generate the mutant vector $\mathbf{v}_r^{g}$ using Eqn.~\ref{eqn:mutation}.
		\STATE Generate the cross vector $\mathbf{u}_r^{g}$ using Eqn.~\ref{eqn:crossover}.
		\STATE $\mathcal{U}^{g} \gets \mathcal{U}^{g} \cup \mathbf{u}_r^{g}$.
		\ENDFOR
        \STATE Evaluate the fitness of each cross vector $\mathbf{u}_r^{g}$ in $\mathcal{U}^{g}$ using OCT fitness \cref{alg:oct_fit}.
        \FOR{$r\in{1,\cdots,N}$}
        \STATE Select the individual with better fitness as $\mathbf{s}_r^{g+1}$ from $\mathbf{s}_{r}^{g}$ and $\mathbf{u}_{r}^{g}$ using \cref{eqn:selection}.
        \STATE $\mathcal{S}^{g+1} \gets \mathcal{S}^{g+1} \cup \mathbf{s}_r^{g+1}$.
        \ENDFOR
		\STATE $\mathbf{s}_{best}^{g+1} \gets \arg \min_{\mathbf{s}} f(\mathbf{s}_r^{g+1}),\; \mathbf{s}_r^{g+1} \in \mathcal{S}^{g+1}$.
        \STATE $g \gets g+1$.
		\ENDWHILE
	\end{algorithmic}
    \end{spacing}
\end{algorithm}

\newpage
\section{Results of hyperparameter tuning with validation sets} \label{apx:tuning}
In this section, we delve into the results of hyperparameter tuning with validation sets, as showcased in \cref{tbl:hyperparameter}. This involved a comparative analysis of our algorithm, \texttt{MH-DEOCT}, with \texttt{CART} and \texttt{IAI-OCT}. It's noteworthy that \texttt{IAI-OCT} is a commercial OCT package developed by LLC Interpretable AI \citep{bertsimasOptimal2017, InterpretableAI}.

In the \texttt{IAI-OCT} package, the complexity parameter $\alpha$ is automatically adjusted using a training-validation-retraining procedure. This function is intrinsic to the package and cannot be disabled. As a result, \texttt{IAI-OCT} was excluded from the comparison in \cref{tbl:comparision}. Instead, we used a self-implemented version named \texttt{LS-OCT} for the comparison. The running time of our implementation is slower than \texttt{IAI-OCT}, likely because \texttt{IAI-OCT} has been fully optimized for speed as a commercial software product. Nevertheless, both \texttt{LS-OCT} and \texttt{IAI-OCT} demonstrate a similar accelerated growth pattern in speed as the sample size increases (as shown in \cref{fig:time_results}). Due to licensing restrictions, \texttt{IAI-OCT} was run on a personal laptop with an AMD 5800H CPU, which has similar sequential performance capabilities to the server used for other tests (in GeekBench 6 for single-core performance, the laptop is scored 1750 and the server is scored 1650).

For our proposed algorithm, \texttt{MH-DEOCT}, we adjusted the complexity parameter $\alpha$ utilizing the same training-validation-retraining procedure as the \texttt{IAI-OCT} package. In comparison, \texttt{CART} requires tuning of an analogous parameter, termed "purity\_thresh", during the pruning operation to efficiently control the complexity of the tree structure. For both \texttt{MH-DEOCT} and \texttt{CART}, we examined 21 parameter values uniformly selected from their parameter domains to identify the optimal parameter value. In terms of the datasets, we divided 65 datasets, up to 1 million samples, into training, validation, and testing subsets following a 50\%-25\%-25\% ratio. This partitioning was repeated ten times, each iteration utilizing a unique random seed. The hyperparameter tuning process was conducted for each partitioned dataset, and the reported results represent the average of all 650 experiments (10 partitions for each of the 65 datasets).

The results in \cref{tbl:hyperparameter} demonstrate that hyperparameter tuning consistently enhances the testing accuracy of all algorithms compared to their non-tuned versions. Specifically, the testing accuracy of the tuned \texttt{MH-DEOCT} improves on average by $1.37\%$ relative to its non-tuned counterpart, albeit at the cost of a slight $1.29\%$ decrease in training accuracy. Moreover, when compared to tuned \texttt{CART} and \texttt{IAI-OCT}, \texttt{MH-DEOCT} exhibits superior performance in testing accuracy by an average of $1.92\%$ and $0.80\%$, respectively. Additionally, \texttt{MH-DEOCT} maintains a higher training accuracy than the other two algorithms.

\begin{table}[!htp]
\caption{Average accuracy of hyperparameter tuning with validation sets on 65 datasets within 1 million samples}
\label{tbl:hyperparameter}
\begin{center}
\renewcommand{\arraystretch}{0.85}
\resizebox{0.85\textwidth}{!}{
\begin{threeparttable}
\begin{tabular}{c|c|ccc|ccc}
\toprule
\multirow{2}{*}{\begin{tabular}[c]{@{}c@{}}Perfor-\\ mance\end{tabular}} & \multirow{2}{*}{Depth} & \multicolumn{3}{c|}{$\alpha=0$} & \multicolumn{3}{c}{With Tuning} \\ \cline{3-8} 
 &  & CART & LS-OCT & MH-DEOCT & CART & IAI-OCT & MH-DEOCT \\ \midrule
\multirow{4}{*}{\begin{tabular}[c]{@{}c@{}}Train\\ (\%)\end{tabular}} & 2 & 79.41 & 81.47 & \textbf{81.64} & 78.93 & 80.77 & 81.25 \\
 & 3 & 84.30 & 86.04 & \textbf{86.82} & 83.30 & 85.54 & {86.26} \\
 & 4 & 87.62 & 88.87 & \textbf{90.24} & 86.00 & 88.21 & {89.19} \\
 & 8 & 95.67 & 96.26 & \textbf{97.30} & 92.11 & 92.84 & {94.57} \\ \midrule
\multirow{4}{*}{\begin{tabular}[c]{@{}c@{}}Test\\ (\%)\end{tabular}} & 2 & 76.91 & 78.19 & 78.21 & 77.43 & 78.13 & \textbf{78.85} \\
 & 3 & 80.45 & 81.57 & 81.78 & 81.16 & 82.10 & \textbf{82.71} \\
 & 4 & 82.42 & 83.32 & 83.60 & 83.22 & 84.23 & \textbf{84.81} \\
 & 8 & 85.78 & 85.80 & 86.31 & 87.03 & 87.54 & \textbf{87.73} \\ \midrule
\multirow{4}{*}{\begin{tabular}[c]{@{}c@{}}Time\\ (s)\end{tabular}} & 2 & 0.03 & 140.34 & 1.51 & 3.25 & 31.18 & 29.46 \\
 & 3 & 0.05 & 231.98 & 3.58 & 2.12 & 69.28 & 69.56 \\
 & 4 & 0.06 & 366.77 & 6.96 & 2.56 & 136.52 & 127.38 \\
 & 8 & 0.12 & 869.29 & 68.77 & 4.05 & 613.60 & 973.11 \\ \bottomrule
\end{tabular}
\end{threeparttable}
}
\end{center}
\vskip -0.1in
\end{table}

\newpage
\section{Comparison with global methods requiring binary datasets} \label{apx:binaryglobal}
We conducted a comprehensive benchmarking of our approach using the state-of-the-art global optimal dynamic programming method, \texttt{GOSDT-Guesses} (2022) [1], which is tailored to binary features. To accommodate continuous features in our evaluation, we employed two binarization techniques: \texttt{MDLP} [2] and \texttt{Guess} [1]. For the \texttt{Guess} binarization, we opted for the most extensive setting from the reference GBDT model in [1] ($n_{est}=50, max_{depth}=2$) to minimize potential data loss during conversion. Our evaluation encompassed 64 datasets, each with fewer than 250,000 samples. The HTSensor dataset (928,991 samples) was excluded due to its lengthy conversion time (over two hours) and the incompatibility of all global methods to process its binarized version. \texttt{MH-DEOCT} employed the same parameters as detailed in our original paper. Notably, \texttt{GOSDT (2022)} failed to generate results for 8 of the 64 binarized datasets. To ensure a fair comparison, we considered only the average results from the remaining 56 datasets, as summarized in \cref{tbl:binarycomparision}.

\begin{table}[htp]
\caption{Comparison of continuous and binary features, dynamic programming and our algorithm on 63 datasets (Continuous datasets have better average train and test accuracy; MH-DEOCT obtains higher accuracy on continuous datasets than GOSDT on binarized datasets.)}
\centering 
\label{tbl:binarycomparision}
\resizebox{0.9\textwidth}{!}{
\begin{threeparttable}
\begin{tabular}{c|cc|cccc}
\hline
\multirow{2}{*}{Performance} & \multirow{2}{*}{\begin{tabular}[c]{@{}c@{}}Dataset\\      Type\end{tabular}} & \multirow{2}{*}{\begin{tabular}[c]{@{}c@{}}\# Features\\      {[}Mean (Max){]}\end{tabular}} & \multicolumn{4}{c}{Depth=2} \\ \cline{4-7} 
 &  &  & CART & \begin{tabular}[c]{@{}c@{}}Quant-BnB\\      (Global)\end{tabular} & \begin{tabular}[c]{@{}c@{}}GOSDT\\      (2022, Global)\end{tabular} & \begin{tabular}[c]{@{}c@{}}MH-DEOCT\\      (Ours)\end{tabular} \\ \hline
\multirow{3}{*}{Train (\%)} & Continuous & 20.1 (73) & 80.68 & \textbf{83.17} & - & \textbf{82.93} \\
 & MDLP-Binarized & 76.1   (1003) & 78.45 & 80.36 & 80.36 & 80.22 \\
 & Guess-Binarized & 30.4 (103) & 79.07 & 82.00 & 82.00 & 81.90 \\ \hline
\multirow{3}{*}{Test (\%)} & Continuous & 20.1 (73) & 77.93 & \textbf{79.43} & - & \textbf{79.42} \\
 & MDLP-Binarized & 76.1   (1003) & 76.20 & 78.29 & 78.28 & 78.11 \\
 & Guess-Binarized & 30.4 (103) & 77.33 & 78.75 & 78.75 & 78.63 \\ \hline
\multirow{3}{*}{Time (s)} & Continuous & 20.1 (73) & 0.01 & 1.76 & - & 1.77 \\
 & MDLP-Binarized & 76.1   (1003) & 0.02 & 49.12 & 6.91 & 3.41 \\
 & Guess-Binarized & 30.4 (103) & 0.01 & 3.23 & 4.33 & 2.11 \\ \hline
\end{tabular}
\begin{tablenotes}
    \tiny
    \item * GOSDT was unable to produce results for 7 out of 64 datasets when binarized using MDLP and for 1 out of 64 datasets when binarized using Guess. For an fair comparison, we present the average results from the 56 out of 64 datasets where GOSDT successfully generated outcomes.
\end{tablenotes}
\end{threeparttable}
}
\end{table}
\vskip -0.1in
\subsection{Comparison of Continuous and Binarized Datasets}
While full binarization methods, such as \texttt{one-hot} encoding, offer consistent search spaces for splits between continuous and corresponding binary datasets, they significantly inflate the feature count. Across the aforementioned 64 datasets, the original average feature count was 20.1, which soared to an average of 4060 post-binarization, with a maximum of 124,968. Handling such vast binary features poses challenges for current global optimal solvers. For context, the largest reported managed binary feature count, 9460 by \texttt{DL8.5} [2], was derived from 19 continuous features with 1151 samples. Furthermore, full binarization leads to substantial increases in storage requirements; for example, a dataset with 17,898 samples grew from 3MB to approximately 9GB after conversion, presenting challenges in addressing real-world large-scale problems.

As a solution to reduce binarized features, various methods have been proposed. \texttt{Guess} [1] derives thresholds from pre-trained boosted decision tree models, \texttt{MDLP} [3] applies the minimum description length principle, and \texttt{Bucketization} [2, 4] avoids splitting between consecutive observations. However, \texttt{Bucketization}'s accuracy loss has been documented by [5]. In our assessments, we evaluated \texttt{MDLP}-binarized, \texttt{Guess}-binarized, and original continuous datasets. As depicted in \cref{tbl:binarycomparision}, \texttt{Quant-BnB} consistently achieves global optima across all three datasets. It's worth noting that the accuracy of the continuous datasets is notably higher compared to both the \texttt{MDLP} and \texttt{Guess} binarized versions, emphasizing the potential information loss when converting numerical features to binary, especially when dataset size management is a priority.

\subsection{Comparison of \texttt{GOSDT-Guesses} and \texttt{MH-DEOCT}}
The comparison presented in \cref{tbl:binarycomparision} underscores that our algorithm achieves results comparable to the global optimal solutions provided by \texttt{GOSDT-Guesses} on binary datasets. Comparing continuous with binarized datasets, \texttt{MH-DEOCT} consistently delivers superior accuracy on continuous datasets in comparison to the optimal results of \texttt{GOSDT-Guess} on the corresponding binarized datasets. Additionally, \cref{tbl:binarydepth} provides results for two datasets with sample sizes exceeding 10,000 over various tree depths. It demonstrates that \texttt{GOSDT-Guesses} encounters challenges in obtaining optimal solutions within a reasonable time for larger datasets and deeper trees (e.g., it failed to produce results within a 4-hour limit for depth=8 in two datasets in the table). In contrast, our algorithm adeptly handles these challenges, efficiently solving problems with larger datasets (up to 11,000,000 samples) and deeper trees, setting a precedent in the literature.

\begin{table}[htp]
\label{tbl:binarydepth}
\caption{Comparison of MH-DEOCT and GOSDT (2022) on Guess-Binarized datasets with different tree depth (GOSDT is slower and failed to solve deep trees and datasets large than 245,057 samples, while our algorithm can solve depth=8 trees with up to 11,000,000 samples)}
\centering
\resizebox{\textwidth}{!}{
\begin{threeparttable}
\begin{tabular}{c|ccc|c|ccc|ccc|ccc}
\hline
\multirow{2}{*}{Dataset} & \multirow{2}{*}{Sample} & \multirow{2}{*}{\begin{tabular}[c]{@{}c@{}}\# Feature\\      (Original/ \\Guess)\end{tabular}} & \multirow{2}{*}{Class} & \multirow{2}{*}{\begin{tabular}[c]{@{}c@{}}Perfor-\\      mance\end{tabular}} & \multicolumn{3}{c|}{Depth = 2} & \multicolumn{3}{c|}{Depth = 4} & \multicolumn{3}{c}{Depth = 8} \\ \cline{6-14} 
 &  &  &  &  & CART & \begin{tabular}[c]{@{}c@{}}GOSDT-\\      Guesses\end{tabular} & MH-DEOCT & CART & \begin{tabular}[c]{@{}c@{}}GOSDT-\\      Guesses\end{tabular} & MH-DEOCT & CART & \begin{tabular}[c]{@{}c@{}}GOSDT-\\      Guesses\end{tabular} & MH-DEOCT \\ \hline
\multirow{3}{*}{Eeg} & \multirow{3}{*}{14,980} & \multirow{3}{*}{14 / 103} & \multirow{3}{*}{2} & Train (\%) & 63.02 & 66.95 & 66.95 & 70.01 & 74.70 & 72.62 & 77.64 & OoT & 78.01 \\
 &  &  &  & Test (\%) & 62.36 & 66.42 & 66.42 & 68.78 & 73.55 & 71.21 & 75.50 & OoT & 75.59 \\
 &  &  &  & Time (s) & 0.03 & 4.64 & 3.73 & 0.07 & 1542 & 21.33 & 0.11 & OoT & 312 \\ \hline
\multirow{3}{*}{\begin{tabular}[c]{@{}c@{}}Skin-\\ seg\end{tabular}} & \multirow{3}{*}{245,057} & \multirow{3}{*}{3 / 63} & \multirow{3}{*}{2} & Train (\%) & 90.71 & 92.70 & 91.97 & 97.43 & 98.16 & 98.06 & 99.29 & OoT & 99.33 \\
 &  &  &  & Test (\%) & 90.67 & 92.60 & 91.89 & 97.38 & 98.18 & 98.02 & 99.27 & OoT & 99.31 \\
 &  &  &  & Time (s) & 0.64 & 90.21 & 11.3 & 0.97 & 1881 & 42.14 & 1.16 & OoT & 508 \\ \hline
\end{tabular}
\begin{tablenotes}
    \item * OOT: Out of 4-hour Time limit.
    \item * For datasets of \textbf{larger sizes} (e.g., ranging from 928,991 to 11,000,000 samples), the binarization process requires over two hours, rendering it unsuitable even as a preprocessing step. However, our algorithm is capable of effectively handling datasets containing up to 11,000,000 samples.
\end{tablenotes}
\end{threeparttable}
}
\end{table}

[1] McTavish, Hayden, et al. "Fast sparse decision tree optimization via reference ensembles." AAAI. 2022.

[2] Demirović, Emir, et al. "Murtree: Optimal decision trees via dynamic programming and search." JMLR 23.1 (2022): 1169-1215.

[3] Verwer, Sicco, and Yingqian Zhang. "Learning optimal classification trees using a binary linear program formulation." AAAI. 2019.

[4] Aglin, Gaël, Siegfried Nijssen, and Pierre Schaus. "Learning optimal decision trees using caching branch-and-bound search." AAAI. 2020.

[5] Lin, Jimmy, et al. "Generalized and scalable optimal sparse decision trees." ICML. 2020.

\section{Detailed experiment results with $\alpha=0$ in \cref{tbl:comparision}} \label{apx:results}
In this section, we expand upon the experiment results presented in \cref{tbl:comparision} by providing a more detailed analysis for each of the 65 datasets. We have documented the average training and testing accuracy, as well as the training time for each dataset based on 10 random splits. These performance metrics offer valuable insight into the stability and efficiency of the models and algorithms employed in our study. To facilitate easier comprehension, the results are presented in a tabular format, with each row corresponding to a dataset and columns representing the average training accuracy, testing accuracy, and training time. By examining these comprehensive results, readers can better understand the strengths and limitations of the methods used, as well as their applicability to various data conditions. 

In these compared methods, \texttt{LS-OCT} is a local-search heuristic method providing node-by-node polishing on the result of CART and other warm-starts \cite{bertsimasOptimal2017}. We use 100 warm-start solutions in \texttt{LS-OCT}, including one CART solution and 99 random generated solutions. It should be noted that \texttt{LS-OCT} was unable to produce a result for the \texttt{Htsensor} dataset, which contains $928,991$ samples, at Depth=8 within a 24-hour timeframe. Therefore, this particular result is not included in the subsequent tables. In addition, the global optimality of the methods \texttt{Quant-BnB} and \texttt{RS-OCT} can be inferred from their training times. With a time limit of four hours for \texttt{Quant-BnB} and \texttt{RS-OCT}, any training time less than these limits indicates that a global optimal solution has been achieved.

\subsection{Training accuracy (\%) on 65 datasets within 1 million samples}
{\tiny\tabcolsep=3pt
\begin{longtable}{cccc|cccccccc}
\caption{Depth=2 average training accuracy (\%) on 65 datasets with 10 random splits within 1 million samples} \\
\hline
\multicolumn{1}{c|}{\multirow{2}{*}{Dataset}} & \multirow{2}{*}{Sample} & \multirow{2}{*}{Feature} & \multirow{2}{*}{Class} & \multicolumn{8}{c}{Depth=2} \\ \cline{5-12} 
\multicolumn{1}{c|}{} &  &  &  & CART & LS-OCT & Quant-BnB & RS-OCT & DEOCT & MH-DEOCT & \begin{tabular}[c]{@{}c@{}}DEOCT\\      (Long)\end{tabular} & \begin{tabular}[c]{@{}c@{}}MH-DEOCT\\      (Long)\end{tabular} \\ \hline
\endfirsthead
\multicolumn{12}{c}%
{{\bfseries Table \thetable\ continued from previous page}} \\
\hline
\multicolumn{1}{c|}{\multirow{2}{*}{Dataset}} & \multirow{2}{*}{Sample} & \multirow{2}{*}{Feature} & \multirow{2}{*}{Class} & \multicolumn{8}{c}{Depth=2} \\ \cline{5-12} 
\multicolumn{1}{c|}{} &  &  &  & CART & LS-OCT & Quant-BnB & RS-OCT & DEOCT & MH-DEOCT & \begin{tabular}[c]{@{}c@{}}DEOCT\\      (Long)\end{tabular} & \begin{tabular}[c]{@{}c@{}}MH-DEOCT\\      (Long)\end{tabular} \\ \hline
\endhead
\hline
\endfoot
\endlastfoot
\multicolumn{1}{c|}{Soybean-small} & 47 & 35 & 4 & 100.00 & 100.00 & 100.00 & 100.00 & 100.00 & 100.00 & 100.00 & 100.00 \\
\multicolumn{1}{c|}{Echocardiogram} & 61 & 11 & 2 & 100.00 & 100.00 & 100.00 & 100.00 & 100.00 & 100.00 & 100.00 & 100.00 \\
\multicolumn{1}{c|}{Hepatitis} & 80 & 19 & 2 & 95.00 & 95.50 & 96.67 & 96.67 & 96.00 & 96.00 & 96.17 & 96.33 \\
\multicolumn{1}{c|}{Fertility} & 100 & 9 & 2 & 87.73 & 89.47 & 90.13 & 90.13 & 90.13 & 90.13 & 90.00 & 90.00 \\
\multicolumn{1}{c|}{Acute-inflammations-1} & 120 & 6 & 2 & 91.44 & 100.00 & 100.00 & 100.00 & 100.00 & 100.00 & 100.00 & 100.00 \\
\multicolumn{1}{c|}{Acute-inflammations-2} & 120 & 6 & 2 & 100.00 & 100.00 & 100.00 & 100.00 & 100.00 & 100.00 & 100.00 & 100.00 \\
\multicolumn{1}{c|}{Hayes–roth} & 132 & 5 & 3 & 58.38 & 62.63 & 63.43 & 63.43 & 63.43 & 63.43 & 63.43 & 63.43 \\
\multicolumn{1}{c|}{Iris} & 150 & 4 & 3 & 96.16 & 96.25 & 96.25 & 96.25 & 96.16 & 96.16 & 96.16 & 96.25 \\
\hline
\end{longtable}
}

{\tiny\tabcolsep=3pt
\begin{longtable}{cccc|cccccccc}
\caption{Depth=2 training accuracy standard deviation (\%) on 65 datasets with 10 random splits within 1 million samples} \\
\hline
\multicolumn{1}{c|}{\multirow{2}{*}{Dataset}} & \multirow{2}{*}{Sample} & \multirow{2}{*}{Feature} & \multirow{2}{*}{Class} & \multicolumn{8}{c}{Depth=2} \\ \cline{5-12} 
\multicolumn{1}{c|}{} &  &  &  & CART & LS-OCT & Quant-BnB & RS-OCT & DEOCT & MH-DEOCT & \begin{tabular}[c]{@{}c@{}}DEOCT\\      (Long)\end{tabular} & \begin{tabular}[c]{@{}c@{}}MH-DEOCT\\      (Long)\end{tabular} \\ \hline
\endfirsthead
\multicolumn{12}{c}%
{{\bfseries Table \thetable\ continued from previous page}} \\
\hline
\multicolumn{1}{c|}{\multirow{2}{*}{Dataset}} & \multirow{2}{*}{Sample} & \multirow{2}{*}{Feature} & \multirow{2}{*}{Class} & \multicolumn{8}{c}{Depth=2} \\ \cline{5-12} 
\multicolumn{1}{c|}{} &  &  &  & CART & LS-OCT & Quant-BnB & RS-OCT & DEOCT & MH-DEOCT & \begin{tabular}[c]{@{}c@{}}DEOCT\\      (Long)\end{tabular} & \begin{tabular}[c]{@{}c@{}}MH-DEOCT\\      (Long)\end{tabular} \\ \hline
\endhead
\hline
\endfoot
\endlastfoot
\multicolumn{1}{c|}{Soybean-small} & 47 & 35 & 4 & 0.00 & 0.00 & 0.00 & 0.00 & 0.00 & 0.00 & 0.00 & 0.00 \\
\multicolumn{1}{c|}{Echocardiogram} & 61 & 11 & 2 & 0.00 & 0.00 & 0.00 & 0.00 & 0.00 & 0.00 & 0.00 & 0.00 \\
\multicolumn{1}{c|}{Hepatitis} & 80 & 19 & 2 & 2.94 & 2.94 & 1.92 & 1.92 & 2.51 & 2.51 & 2.49 & 2.33 \\
\multicolumn{1}{c|}{Fertility} & 100 & 9 & 2 & 2.33 & 1.93 & 1.43 & 1.43 & 1.43 & 1.43 & 1.69 & 1.69 \\
\multicolumn{1}{c|}{Acute-inflammations-1} & 120 & 6 & 2 & 1.18 & 0.00 & 0.00 & 0.00 & 0.00 & 0.00 & 0.00 & 0.00 \\
\multicolumn{1}{c|}{Acute-inflammations-2} & 120 & 6 & 2 & 0.00 & 0.00 & 0.00 & 0.00 & 0.00 & 0.00 & 0.00 & 0.00 \\
\multicolumn{1}{c|}{Hayes–roth} & 132 & 5 & 3 & 1.89 & 2.13 & 1.56 & 1.56 & 1.56 & 1.56 & 1.56 & 1.56 \\
\multicolumn{1}{c|}{Iris} & 150 & 4 & 3 & 1.04 & 1.18 & 1.18 & 1.18 & 1.04 & 1.04 & 1.04 & 1.18 \\
\hline
\end{longtable}
}

{\tiny\tabcolsep=3pt
\begin{longtable}{cccc|ccccccc}
\caption{Depth=3 average training accuracy (\%) on 65 datasets with 10 random splits within 1 million samples} \\
\hline
\multicolumn{1}{c|}{\multirow{2}{*}{Dataset}} & \multirow{2}{*}{Sample} & \multirow{2}{*}{Feature} & \multirow{2}{*}{Class} & \multicolumn{7}{c}{Depth=3} \\ \cline{5-11} 
\multicolumn{1}{c|}{} &  &  &  & CART & LS-OCT & Quant-BnB & DEOCT & MH-DEOCT & \begin{tabular}[c]{@{}c@{}}DEOCT\\      (Long)\end{tabular} & \begin{tabular}[c]{@{}c@{}}MH-DEOCT\\      (Long)\end{tabular} \\ \hline
\endfirsthead
\multicolumn{11}{c}%
{{\bfseries Table \thetable\ continued from previous page}} \\
\hline
\multicolumn{1}{c|}{\multirow{2}{*}{Dataset}} & \multirow{2}{*}{Sample} & \multirow{2}{*}{Feature} & \multirow{2}{*}{Class} & \multicolumn{7}{c}{Depth=3} \\ \cline{5-11} 
\multicolumn{1}{c|}{} &  &  &  & CART & LS-OCT & Quant-BnB & DEOCT & MH-DEOCT & \begin{tabular}[c]{@{}c@{}}DEOCT\\      (Long)\end{tabular} & \begin{tabular}[c]{@{}c@{}}MH-DEOCT\\      (Long)\end{tabular} \\ \hline
\endhead
\hline
\endfoot
\endlastfoot
\multicolumn{1}{c|}{Soybean-small} & 47 & 35 & 4 & 100.00 & 100.00 & - & 100.00 & 100.00 & 100.00 & 100.00 \\
\multicolumn{1}{c|}{Echocardiogram} & 61 & 11 & 2 & 100.00 & 100.00 & 100.00 & 100.00 & 100.00 & 100.00 & 100.00 \\
\multicolumn{1}{c|}{Hepatitis} & 80 & 19 & 2 & 97.50 & 97.83 & 100.00 & 98.33 & 99.00 & 99.33 & 99.33 \\
\multicolumn{1}{c|}{Fertility} & 100 & 9 & 2 & 90.40 & 91.73 & 93.87 & 92.80 & 92.93 & 93.73 & 94.27 \\
\multicolumn{1}{c|}{Acute-inflammations-1} & 120 & 6 & 2 & 100.00 & 100.00 & 100.00 & 100.00 & 100.00 & 100.00 & 100.00 \\
\multicolumn{1}{c|}{Acute-inflammations-2} & 120 & 6 & 2 & 100.00 & 100.00 & 100.00 & 100.00 & 100.00 & 100.00 & 100.00 \\
\multicolumn{1}{c|}{Hayes–roth} & 132 & 5 & 3 & 64.04 & 70.61 & 75.56 & 72.53 & 72.83 & 73.43 & 74.04 \\
\multicolumn{1}{c|}{Iris} & 150 & 4 & 3 & 97.14 & 98.21 & 99.82 & 98.21 & 98.21 & 98.75 & 99.20 \\
\hline
\end{longtable}
}

{\tiny\tabcolsep=3pt

\begin{longtable}{cccc|ccccccc}
\caption{Depth=3 training accuracy standard deviation (\%) on 65 datasets with 10 random splits within 1 million samples} \\
\hline
\multicolumn{1}{c|}{\multirow{2}{*}{Dataset}} & \multirow{2}{*}{Sample} & \multirow{2}{*}{Feature} & \multirow{2}{*}{Class} & \multicolumn{7}{c}{Depth=3} \\ \cline{5-11} 
\multicolumn{1}{c|}{} &  &  &  & CART & LS-OCT & Quant-BnB & DEOCT & MH-DEOCT & \begin{tabular}[c]{@{}c@{}}DEOCT\\      (Long)\end{tabular} & \begin{tabular}[c]{@{}c@{}}MH-DEOCT\\      (Long)\end{tabular} \\ \hline
\endfirsthead
\multicolumn{11}{c}%
{{\bfseries Table \thetable\ continued from previous page}} \\
\hline
\multicolumn{1}{c|}{\multirow{2}{*}{Dataset}} & \multirow{2}{*}{Sample} & \multirow{2}{*}{Feature} & \multirow{2}{*}{Class} & \multicolumn{7}{c}{Depth=3} \\ \cline{5-11} 
\multicolumn{1}{c|}{} &  &  &  & CART & LS-OCT & Quant-BnB & DEOCT & MH-DEOCT & \begin{tabular}[c]{@{}c@{}}DEOCT\\      (Long)\end{tabular} & \begin{tabular}[c]{@{}c@{}}MH-DEOCT\\      (Long)\end{tabular} \\ \hline
\endhead
\hline
\endfoot
\endlastfoot
\multicolumn{1}{c|}{Soybean-small} & 47 & 35 & 4 & 0.00 & 0.00 & - & 0.00 & 0.00 & 0.00 & 0.00 \\
\multicolumn{1}{c|}{Echocardiogram} & 61 & 11 & 2 & 0.00 & 0.00 & 0.00 & 0.00 & 0.00 & 0.00 & 0.00 \\
\multicolumn{1}{c|}{Hepatitis} & 80 & 19 & 2 & 2.39 & 1.93 & 0.00 & 1.57 & 1.17 & 0.86 & 0.86 \\
\multicolumn{1}{c|}{Fertility} & 100 & 9 & 2 & 2.87 & 1.64 & 2.23 & 1.43 & 1.26 & 1.78 & 1.99 \\
\multicolumn{1}{c|}{Acute-inflammations-1} & 120 & 6 & 2 & 0.00 & 0.00 & 0.00 & 0.00 & 0.00 & 0.00 & 0.00 \\
\multicolumn{1}{c|}{Acute-inflammations-2} & 120 & 6 & 2 & 0.00 & 0.00 & 0.00 & 0.00 & 0.00 & 0.00 & 0.00 \\
\multicolumn{1}{c|}{Hayes–roth} & 132 & 5 & 3 & 1.79 & 2.49 & 2.19 & 2.60 & 2.75 & 2.24 & 2.02 \\
\multicolumn{1}{c|}{Iris} & 150 & 4 & 3 & 0.82 & 0.60 & 0.40 & 0.60 & 0.60 & 0.96 & 0.89 \\
\hline
\end{longtable}
}

{\tiny\tabcolsep=3pt
\setlength{\LTleft}{-20cm plus -1fill}
\setlength{\LTright}{\LTleft}
\begin{longtable}{c|ccc|cccccc|cccccc}
\caption{Depth=4,8 average training accuracy (\%) on 65 datasets with 10 random splits within 1 million samples} \\
\hline
\multicolumn{1}{c|}{\multirow{2}{*}{Dataset}} & \multirow{2}{*}{Sample} & \multirow{2}{*}{Feature} & \multirow{2}{*}{Class} & \multicolumn{6}{c|}{Depth=4} & \multicolumn{6}{c}{Depth=8} \\ \cline{5-16} 
\multicolumn{1}{c|}{} &  &  &  & CART & LS-OCT & DEOCT & MH-DEOCT & \begin{tabular}[c]{@{}c@{}}DEOCT\\      (Long)\end{tabular} & \begin{tabular}[c]{@{}c@{}}MH-DEOCT\\      (Long)\end{tabular} & CART & LS-OCT & DEOCT & MH-DEOCT & \begin{tabular}[c]{@{}c@{}}DEOCT\\      (Long)\end{tabular} & \begin{tabular}[c]{@{}c@{}}MH-DEOCT\\      (Long)\end{tabular} \\ \hline
\endfirsthead
\multicolumn{16}{c}%
{{\bfseries Table \thetable\ continued from previous page}} \\
\hline
\multicolumn{1}{c|}{\multirow{2}{*}{Dataset}} & \multirow{2}{*}{Sample} & \multirow{2}{*}{Feature} & \multirow{2}{*}{Class} & \multicolumn{6}{c|}{Depth=4} & \multicolumn{6}{c}{Depth=8} \\ \cline{5-16} 
\multicolumn{1}{c|}{} &  &  &  & CART & LS-OCT & DEOCT & MH-DEOCT & \begin{tabular}[c]{@{}c@{}}DEOCT\\      (Long)\end{tabular} & \begin{tabular}[c]{@{}c@{}}MH-DEOCT\\      (Long)\end{tabular} & CART & LS-OCT & DEOCT & MH-DEOCT & \begin{tabular}[c]{@{}c@{}}DEOCT\\      (Long)\end{tabular} & \begin{tabular}[c]{@{}c@{}}MH-DEOCT\\      (Long)\end{tabular} \\ \hline
\endhead
\hline
\endfoot
\endlastfoot
\multicolumn{1}{c|}{Soybean-small} & 47 & 35 & 4 & 100.00 & 100.00 & 100.00 & 100.00 & 100.00 & 100.00 & 100.00 & 100.00 & 100.00 & 100.00 & 100.00 & 100.00 \\
\multicolumn{1}{c|}{Echocardiogram} & 61 & 11 & 2 & 100.00 & 100.00 & 100.00 & 100.00 & 100.00 & 100.00 & 100.00 & 100.00 & 100.00 & 100.00 & 100.00 & 100.00 \\
\multicolumn{1}{c|}{Hepatitis} & 80 & 19 & 2 & 99.50 & 99.50 & 99.50 & 99.83 & 100.00 & 100.00 & 100.00 & 100.00 & 100.00 & 100.00 & 100.00 & 100.00 \\
\multicolumn{1}{c|}{Fertility} & 100 & 9 & 2 & 93.60 & 94.13 & 94.93 & 96.13 & 96.53 & 97.20 & 98.67 & 99.20 & 98.89 & 98.89 & 98.93 & 98.93 \\
\multicolumn{1}{c|}{Acute-inflammations-1} & 120 & 6 & 2 & 100.00 & 100.00 & 100.00 & 100.00 & 100.00 & 100.00 & 100.00 & 100.00 & 100.00 & 100.00 & 100.00 & 100.00 \\
\multicolumn{1}{c|}{Acute-inflammations-2} & 120 & 6 & 2 & 100.00 & 100.00 & 100.00 & 100.00 & 100.00 & 100.00 & 100.00 & 100.00 & 100.00 & 100.00 & 100.00 & 100.00 \\
\multicolumn{1}{c|}{Hayes–roth} & 132 & 5 & 3 & 74.24 & 76.57 & 78.69 & 81.82 & 84.75 & 86.16 & 93.94 & 95.56 & 94.28 & 96.97 & 96.57 & 98.79 \\
\multicolumn{1}{c|}{Iris} & 150 & 4 & 3 & 99.46 & 99.64 & 99.82 & 99.82 & 99.82 & 99.91 & 100.00 & 100.00 & 100.00 & 100.00 & 100.00 & 100.00 \\
\hline
\end{longtable}
}

{\tiny\tabcolsep=3pt
\setlength{\LTleft}{-20cm plus -1fill}
\setlength{\LTright}{\LTleft}
\begin{longtable}{cccc|cccccc|cccccc}
\caption{Depth=4,8 training accuracy standard deviation (\%) on 65 datasets with 10 splits within 1 million samples} \\
\hline
\multicolumn{1}{c|}{\multirow{2}{*}{Dataset}} & \multirow{2}{*}{Sample} & \multirow{2}{*}{Feature} & \multirow{2}{*}{Class} & \multicolumn{6}{c|}{Depth=4} & \multicolumn{6}{c}{Depth=8} \\ \cline{5-16} 
\multicolumn{1}{c|}{} &  &  &  & CART & LS-OCT & DEOCT & MH-DEOCT & \begin{tabular}[c]{@{}c@{}}DEOCT\\      (Long)\end{tabular} & \begin{tabular}[c]{@{}c@{}}MH-DEOCT\\      (Long)\end{tabular} & CART & LS-OCT & DEOCT & MH-DEOCT & \begin{tabular}[c]{@{}c@{}}DEOCT\\      (Long)\end{tabular} & \begin{tabular}[c]{@{}c@{}}MH-DEOCT\\      (Long)\end{tabular} \\ \hline
\endfirsthead
\multicolumn{16}{c}%
{{\bfseries Table \thetable\ continued from previous page}} \\
\hline
\multicolumn{1}{c|}{\multirow{2}{*}{Dataset}} & \multirow{2}{*}{Sample} & \multirow{2}{*}{Feature} & \multirow{2}{*}{Class} & \multicolumn{6}{c|}{Depth=4} & \multicolumn{6}{c}{Depth=8} \\ \cline{5-16} 
\multicolumn{1}{c|}{} &  &  &  & CART & LS-OCT & DEOCT & MH-DEOCT & \begin{tabular}[c]{@{}c@{}}DEOCT\\      (Long)\end{tabular} & \begin{tabular}[c]{@{}c@{}}MH-DEOCT\\      (Long)\end{tabular} & CART & LS-OCT & DEOCT & MH-DEOCT & \begin{tabular}[c]{@{}c@{}}DEOCT\\      (Long)\end{tabular} & \begin{tabular}[c]{@{}c@{}}MH-DEOCT\\      (Long)\end{tabular} \\ \hline
\endhead
\hline
\endfoot
\endlastfoot
\multicolumn{1}{c|}{Soybean-small} & 47 & 35 & 4 & 0.00 & 0.00 & 0.00 & 0.00 & 0.00 & 0.00 & 0.00 & 0.00 & 0.00 & 0.00 & 0.00 & 0.00 \\
\multicolumn{1}{c|}{Echocardiogram} & 61 & 11 & 2 & 0.00 & 0.00 & 0.00 & 0.00 & 0.00 & 0.00 & 0.00 & 0.00 & 0.00 & 0.00 & 0.00 & 0.00 \\
\multicolumn{1}{c|}{Hepatitis} & 80 & 19 & 2 & 0.81 & 0.76 & 0.81 & 0.53 & 0.00 & 0.00 & 0.00 & 0.00 & 0.00 & 0.00 & 0.00 & 0.00 \\
\multicolumn{1}{c|}{Fertility} & 100 & 9 & 2 & 2.16 & 1.71 & 1.51 & 1.47 & 1.43 & 1.17 & 1.17 & 0.69 & 1.00 & 0.92 & 0.60 & 0.60 \\
\multicolumn{1}{c|}{Acute-inflammations-1} & 120 & 6 & 2 & 0.00 & 0.00 & 0.00 & 0.00 & 0.00 & 0.00 & 0.00 & 0.00 & 0.00 & 0.00 & 0.00 & 0.00 \\
\multicolumn{1}{c|}{Acute-inflammations-2} & 120 & 6 & 2 & 0.00 & 0.00 & 0.00 & 0.00 & 0.00 & 0.00 & 0.00 & 0.00 & 0.00 & 0.00 & 0.00 & 0.00 \\
\multicolumn{1}{c|}{Hayes–roth} & 132 & 5 & 3 & 1.45 & 2.66 & 2.79 & 3.63 & 2.45 & 2.13 & 1.64 & 1.09 & 2.09 & 1.01 & 0.90 & 1.32 \\
\multicolumn{1}{c|}{Iris} & 150 & 4 & 3 & 0.62 & 0.44 & 0.38 & 0.38 & 0.38 & 0.28 & 0.00 & 0.00 & 0.00 & 0.00 & 0.00 & 0.00 \\
\hline
\end{longtable}
}

\subsection{Testing accuracy (\%) on 65 datasets within 1 million samples}
{\tiny\tabcolsep=3pt
\begin{longtable}{cccc|cccccccc}
\caption{Depth=2 average testing accuracy (\%) on 65 datasets with 10 random splits within 1 million samples} \\
\hline
\multicolumn{1}{c|}{\multirow{2}{*}{Dataset}} & \multirow{2}{*}{Sample} & \multirow{2}{*}{Feature} & \multirow{2}{*}{Class} & \multicolumn{8}{c}{Depth=2} \\ \cline{5-12} 
\multicolumn{1}{c|}{} &  &  &  & CART & LS-OCT & Quant-BnB & RS-OCT & DEOCT & MH-DEOCT & \begin{tabular}[c]{@{}c@{}}DEOCT\\      (Long)\end{tabular} & \begin{tabular}[c]{@{}c@{}}MH-DEOCT\\      (Long)\end{tabular} \\ \hline
\endfirsthead
\multicolumn{12}{c}%
{{\bfseries Table \thetable\ continued from previous page}} \\
\hline
\multicolumn{1}{c|}{\multirow{2}{*}{Dataset}} & \multirow{2}{*}{Sample} & \multirow{2}{*}{Feature} & \multirow{2}{*}{Class} & \multicolumn{8}{c}{Depth=2} \\ \cline{5-12} 
\multicolumn{1}{c|}{} &  &  &  & CART & LS-OCT & Quant-BnB & RS-OCT & DEOCT & MH-DEOCT & \begin{tabular}[c]{@{}c@{}}DEOCT\\      (Long)\end{tabular} & \begin{tabular}[c]{@{}c@{}}MH-DEOCT\\      (Long)\end{tabular} \\ \hline
\endhead
\hline
\endfoot
\endlastfoot
\multicolumn{1}{c|}{Soybean-small} & 47 & 35 & 4 & 98.33 & 98.33 & 98.33 & 100.00 & 98.33 & 100.00 & 98.33 & 100.00 \\
\multicolumn{1}{c|}{Echocardiogram} & 61 & 11 & 2 & 96.25 & 96.25 & 97.50 & 97.50 & 95.63 & 98.13 & 95.63 & 97.50 \\
\multicolumn{1}{c|}{Hepatitis} & 80 & 19 & 2 & 81.00 & 80.50 & 78.00 & 81.00 & 80.50 & 79.50 & 80.00 & 78.50 \\
\multicolumn{1}{c|}{Fertility} & 100 & 9 & 2 & 88.80 & 84.00 & 84.40 & 85.60 & 86.00 & 86.40 & 85.20 & 86.00 \\
\multicolumn{1}{c|}{Acute-inflammations-1} & 120 & 6 & 2 & 92.33 & 100.00 & 100.00 & 100.00 & 100.00 & 100.00 & 100.00 & 100.00 \\
\multicolumn{1}{c|}{Acute-inflammations-2} & 120 & 6 & 2 & 100.00 & 100.00 & 99.67 & 100.00 & 100.00 & 100.00 & 100.00 & 100.00 \\
\multicolumn{1}{c|}{Hayes–roth} & 132 & 5 & 3 & 46.06 & 46.06 & 51.21 & 51.21 & 50.30 & 50.30 & 50.61 & 50.61 \\
\multicolumn{1}{c|}{Iris} & 150 & 4 & 3 & 95.00 & 94.47 & 93.95 & 93.95 & 95.00 & 94.47 & 95.00 & 93.95 \\
\hline
\end{longtable}
}

{\tiny\tabcolsep=3pt
\begin{longtable}{cccc|cccccccc}
\caption{Depth=2 testing accuracy standard deviation (\%) on 65 datasets with 10 random splits within 1 million samples} \\
\hline
\multicolumn{1}{c|}{\multirow{2}{*}{Dataset}} & \multirow{2}{*}{Sample} & \multirow{2}{*}{Feature} & \multirow{2}{*}{Class} & \multicolumn{8}{c}{Depth=2} \\ \cline{5-12} 
\multicolumn{1}{c|}{} &  &  &  & CART & LS-OCT & Quant-BnB & RS-OCT & DEOCT & MH-DEOCT & \begin{tabular}[c]{@{}c@{}}DEOCT\\      (Long)\end{tabular} & \begin{tabular}[c]{@{}c@{}}MH-DEOCT\\      (Long)\end{tabular} \\ \hline
\endfirsthead
\multicolumn{12}{c}%
{{\bfseries Table \thetable\ continued from previous page}} \\
\hline
\multicolumn{1}{c|}{\multirow{2}{*}{Dataset}} & \multirow{2}{*}{Sample} & \multirow{2}{*}{Feature} & \multirow{2}{*}{Class} & \multicolumn{8}{c}{Depth=2} \\ \cline{5-12} 
\multicolumn{1}{c|}{} &  &  &  & CART & LS-OCT & Quant-BnB & RS-OCT & DEOCT & MH-DEOCT & \begin{tabular}[c]{@{}c@{}}DEOCT\\      (Long)\end{tabular} & \begin{tabular}[c]{@{}c@{}}MH-DEOCT\\      (Long)\end{tabular} \\ \hline
\endhead
\hline
\endfoot
\endlastfoot
\multicolumn{1}{c|}{Soybean-small} & 47 & 35 & 4 & 3.51 & 3.51 & 5.27 & 0.00 & 5.27 & 0.00 & 5.27 & 0.00 \\
\multicolumn{1}{c|}{Echocardiogram} & 61 & 11 & 2 & 3.23 & 3.23 & 3.23 & 3.29 & 3.02 & 3.02 & 3.02 & 3.23 \\
\multicolumn{1}{c|}{Hepatitis} & 80 & 19 & 2 & 7.75 & 8.32 & 7.89 & 8.10 & 8.96 & 8.64 & 6.67 & 9.14 \\
\multicolumn{1}{c|}{Fertility} & 100 & 9 & 2 & 7.00 & 9.61 & 7.41 & 6.31 & 7.83 & 7.11 & 6.81 & 7.60 \\
\multicolumn{1}{c|}{Acute-inflammations-1} & 120 & 6 & 2 & 3.53 & 0.00 & 0.00 & 0.00 & 0.00 & 0.00 & 0.00 & 0.00 \\
\multicolumn{1}{c|}{Acute-inflammations-2} & 120 & 6 & 2 & 0.00 & 0.00 & 1.05 & 0.00 & 0.00 & 0.00 & 0.00 & 0.00 \\
\multicolumn{1}{c|}{Hayes–roth} & 132 & 5 & 3 & 5.68 & 9.24 & 5.61 & 8.90 & 6.73 & 6.73 & 5.35 & 5.35 \\
\multicolumn{1}{c|}{Iris} & 150 & 4 & 3 & 3.15 & 4.01 & 3.73 & 3.34 & 3.15 & 2.90 & 3.15 & 3.73 \\
\hline
\end{longtable}
}

{\tiny\tabcolsep=3pt
\begin{longtable}{cccc|ccccccc}
\caption{Depth=3 average testing accuracy (\%) on 65 datasets with 10 random splits within 1 million samples} \\
\hline
\multicolumn{1}{c|}{\multirow{2}{*}{Dataset}} & \multirow{2}{*}{Sample} & \multirow{2}{*}{Feature} & \multirow{2}{*}{Class} & \multicolumn{7}{c}{Depth=3} \\ \cline{5-11} 
\multicolumn{1}{c|}{} &  &  &  & CART & LS-OCT & Quant-BnB & DEOCT & MH-DEOCT & \begin{tabular}[c]{@{}c@{}}DEOCT\\      (Long)\end{tabular} & \begin{tabular}[c]{@{}c@{}}MH-DEOCT\\      (Long)\end{tabular} \\ \hline
\endfirsthead
\multicolumn{11}{c}%
{{\bfseries Table \thetable\ continued from previous page}} \\
\hline
\multicolumn{1}{c|}{\multirow{2}{*}{Dataset}} & \multirow{2}{*}{Sample} & \multirow{2}{*}{Feature} & \multirow{2}{*}{Class} & \multicolumn{7}{c}{Depth=3} \\ \cline{5-11} 
\multicolumn{1}{c|}{} &  &  &  & CART & LS-OCT & Quant-BnB & DEOCT & MH-DEOCT & \begin{tabular}[c]{@{}c@{}}DEOCT\\      (Long)\end{tabular} & \begin{tabular}[c]{@{}c@{}}MH-DEOCT\\      (Long)\end{tabular} \\ \hline
\endhead
\hline
\endfoot
\endlastfoot
\multicolumn{1}{c|}{Soybean-small} & 47 & 35 & 4 & 98.33 & 98.33 & - & 98.33 & 100.00 & 98.33 & 100.00 \\
\multicolumn{1}{c|}{Echocardiogram} & 61 & 11 & 2 & 96.25 & 96.25 & 100.00 & 98.13 & 96.25 & 96.88 & 98.13 \\
\multicolumn{1}{c|}{Hepatitis} & 80 & 19 & 2 & 81.00 & 78.00 & 80.00 & 79.50 & 81.00 & 80.00 & 80.00 \\
\multicolumn{1}{c|}{Fertility} & 100 & 9 & 2 & 82.80 & 82.80 & 80.80 & 80.80 & 81.20 & 82.40 & 80.40 \\
\multicolumn{1}{c|}{Acute-inflammations-1} & 120 & 6 & 2 & 99.33 & 99.33 & 100.00 & 100.00 & 99.00 & 100.00 & 100.00 \\
\multicolumn{1}{c|}{Acute-inflammations-2} & 120 & 6 & 2 & 100.00 & 100.00 & 100.00 & 100.00 & 100.00 & 100.00 & 99.67 \\
\multicolumn{1}{c|}{Hayes–roth} & 132 & 5 & 3 & 53.33 & 50.91 & 52.73 & 53.94 & 54.55 & 53.64 & 52.42 \\
\multicolumn{1}{c|}{Iris} & 150 & 4 & 3 & 95.79 & 95.79 & 95.26 & 96.32 & 96.05 & 95.79 & 96.05 \\
\hline
\end{longtable}
}

{\tiny\tabcolsep=3pt
\begin{longtable}{cccc|ccccccc}
\caption{Depth=3 testing accuracy standard deviation (\%) on 65 datasets with 10 random splits within 1 million samples} \\
\hline
\multicolumn{1}{c|}{\multirow{2}{*}{Dataset}} & \multirow{2}{*}{Sample} & \multirow{2}{*}{Feature} & \multirow{2}{*}{Class} & \multicolumn{7}{c}{Depth=3} \\ \cline{5-11} 
\multicolumn{1}{c|}{} &  &  &  & CART & LS-OCT & Quant-BnB & DEOCT & MH-DEOCT & \begin{tabular}[c]{@{}c@{}}DEOCT\\      (Long)\end{tabular} & \begin{tabular}[c]{@{}c@{}}MH-DEOCT\\      (Long)\end{tabular} \\ \hline
\endfirsthead
\multicolumn{11}{c}%
{{\bfseries Table \thetable\ continued from previous page}} \\
\hline
\multicolumn{1}{c|}{\multirow{2}{*}{Dataset}} & \multirow{2}{*}{Sample} & \multirow{2}{*}{Feature} & \multirow{2}{*}{Class} & \multicolumn{7}{c}{Depth=3} \\ \cline{5-11} 
\multicolumn{1}{c|}{} &  &  &  & CART & LS-OCT & Quant-BnB & DEOCT & MH-DEOCT & \begin{tabular}[c]{@{}c@{}}DEOCT\\      (Long)\end{tabular} & \begin{tabular}[c]{@{}c@{}}MH-DEOCT\\      (Long)\end{tabular} \\ \hline
\endhead
\hline
\endfoot
\endlastfoot
\multicolumn{1}{c|}{Soybean-small} & 47 & 35 & 4 & 3.51 & 3.51 & - & 5.27 & 0.00 & 5.27 & 0.00 \\
\multicolumn{1}{c|}{Echocardiogram} & 61 & 11 & 2 & 3.23 & 3.23 & 0.00 & 3.02 & 6.04 & 4.42 & 3.02 \\
\multicolumn{1}{c|}{Hepatitis} & 80 & 19 & 2 & 7.38 & 8.56 & 6.12 & 7.25 & 4.59 & 5.77 & 4.71 \\
\multicolumn{1}{c|}{Fertility} & 100 & 9 & 2 & 8.65 & 5.35 & 9.96 & 10.12 & 8.01 & 9.28 & 8.10 \\
\multicolumn{1}{c|}{Acute-inflammations-1} & 120 & 6 & 2 & 2.11 & 2.11 & 0.00 & 0.00 & 3.16 & 0.00 & 0.00 \\
\multicolumn{1}{c|}{Acute-inflammations-2} & 120 & 6 & 2 & 0.00 & 0.00 & 0.00 & 0.00 & 0.00 & 0.00 & 1.05 \\
\multicolumn{1}{c|}{Hayes–roth} & 132 & 5 & 3 & 5.38 & 10.08 & 11.26 & 7.67 & 8.08 & 9.48 & 9.04 \\
\multicolumn{1}{c|}{Iris} & 150 & 4 & 3 & 3.33 & 3.96 & 4.32 & 2.83 & 3.77 & 3.96 & 3.97 \\
\hline
\end{longtable}
}

{\tiny\tabcolsep=3pt
\setlength{\LTleft}{-20cm plus -1fill}
\setlength{\LTright}{\LTleft}
\begin{longtable}{cccc|cccccc|cccccc}
\caption{Depth=4,8 average testing accuracy (\%) on 65 datasets with 10 random splits within 1 million samples} \\
\hline
\multicolumn{1}{c|}{\multirow{2}{*}{Dataset}} & \multirow{2}{*}{Sample} & \multirow{2}{*}{Feature} & \multirow{2}{*}{Class} & \multicolumn{6}{c|}{Depth=4} & \multicolumn{6}{c}{Depth=8} \\ \cline{5-16} 
\multicolumn{1}{c|}{} &  &  &  & CART & LS-OCT & DEOCT & MH-DEOCT & \begin{tabular}[c]{@{}c@{}}DEOCT\\      (Long)\end{tabular} & \begin{tabular}[c]{@{}c@{}}MH-DEOCT\\      (Long)\end{tabular} & CART & LS-OCT & DEOCT & MH-DEOCT & \begin{tabular}[c]{@{}c@{}}DEOCT\\      (Long)\end{tabular} & \begin{tabular}[c]{@{}c@{}}MH-DEOCT\\      (Long)\end{tabular} \\ \hline
\endfirsthead
\multicolumn{16}{c}%
{{\bfseries Table \thetable\ continued from previous page}} \\
\hline
\multicolumn{1}{c|}{\multirow{2}{*}{Dataset}} & \multirow{2}{*}{Sample} & \multirow{2}{*}{Feature} & \multirow{2}{*}{Class} & \multicolumn{6}{c|}{Depth=4} & \multicolumn{6}{c}{Depth=8} \\ \cline{5-16} 
\multicolumn{1}{c|}{} &  &  &  & CART & LS-OCT & DEOCT & MH-DEOCT & \begin{tabular}[c]{@{}c@{}}DEOCT\\      (Long)\end{tabular} & \begin{tabular}[c]{@{}c@{}}MH-DEOCT\\      (Long)\end{tabular} & CART & LS-OCT & DEOCT & MH-DEOCT & \begin{tabular}[c]{@{}c@{}}DEOCT\\      (Long)\end{tabular} & \begin{tabular}[c]{@{}c@{}}MH-DEOCT\\      (Long)\end{tabular} \\ \hline
\endhead
\hline
\endfoot
\endlastfoot
\multicolumn{1}{c|}{Soybean-small} & 47 & 35 & 4 & 98.33 & 98.33 & 98.33 & 98.33 & 98.33 & 100.00 & 98.33 & 98.33 & 95.83 & 100.00 & 93.33 & 100.00 \\
\multicolumn{1}{c|}{Echocardiogram} & 61 & 11 & 2 & 96.25 & 96.25 & 96.25 & 96.25 & 95.00 & 96.25 & 96.25 & 96.25 & 87.50 & 93.75 & 85.00 & 95.00 \\
\multicolumn{1}{c|}{Hepatitis} & 80 & 19 & 2 & 81.00 & 81.00 & 81.00 & 82.50 & 81.00 & 79.00 & 81.50 & 81.50 & 80.00 & 80.83 & 80.00 & 83.00 \\
\multicolumn{1}{c|}{Fertility} & 100 & 9 & 2 & 84.40 & 84.40 & 83.60 & 80.40 & 82.40 & 80.80 & 79.20 & 74.80 & 82.67 & 84.00 & 80.00 & 87.20 \\
\multicolumn{1}{c|}{Acute-inflammations-1} & 120 & 6 & 2 & 99.33 & 99.33 & 100.00 & 100.00 & 100.00 & 100.00 & 99.33 & 99.33 & 100.00 & 100.00 & 100.00 & 100.00 \\
\multicolumn{1}{c|}{Acute-inflammations-2} & 120 & 6 & 2 & 100.00 & 100.00 & 100.00 & 100.00 & 100.00 & 100.00 & 100.00 & 100.00 & 100.00 & 100.00 & 100.00 & 100.00 \\
\multicolumn{1}{c|}{Hayes–roth} & 132 & 5 & 3 & 61.82 & 60.00 & 58.18 & 59.39 & 65.76 & 65.45 & 78.79 & 69.70 & 76.77 & 69.19 & 72.12 & 65.45 \\
\multicolumn{1}{c|}{Iris} & 150 & 4 & 3 & 96.05 & 95.00 & 95.53 & 95.53 & 95.53 & 96.32 & 96.05 & 96.05 & 96.49 & 95.61 & 95.79 & 97.89 \\
\hline
\end{longtable}
}

{\tiny\tabcolsep=3pt
\setlength{\LTleft}{-20cm plus -1fill}
\setlength{\LTright}{\LTleft}
\begin{longtable}{cccc|cccccc|cccccc}
\caption{Depth=4,8 testing accuracy standard deviation (\%) on 65 datasets with 10 random splits within 1 million samples} \\
\hline
\multicolumn{1}{c|}{\multirow{2}{*}{Dataset}} & \multirow{2}{*}{Sample} & \multirow{2}{*}{Feature} & \multirow{2}{*}{Class} & \multicolumn{6}{c|}{Depth=4} & \multicolumn{6}{c}{Depth=8} \\ \cline{5-16} 
\multicolumn{1}{c|}{} &  &  &  & CART & LS-OCT & DEOCT & MH-DEOCT & \begin{tabular}[c]{@{}c@{}}DEOCT\\      (Long)\end{tabular} & \begin{tabular}[c]{@{}c@{}}MH-DEOCT\\      (Long)\end{tabular} & CART & LS-OCT & DEOCT & MH-DEOCT & \begin{tabular}[c]{@{}c@{}}DEOCT\\      (Long)\end{tabular} & \begin{tabular}[c]{@{}c@{}}MH-DEOCT\\      (Long)\end{tabular} \\ \hline
\endfirsthead
\multicolumn{16}{c}%
{{\bfseries Table \thetable\ continued from previous page}} \\
\hline
\multicolumn{1}{c|}{\multirow{2}{*}{Dataset}} & \multirow{2}{*}{Sample} & \multirow{2}{*}{Feature} & \multirow{2}{*}{Class} & \multicolumn{6}{c|}{Depth=4} & \multicolumn{6}{c}{Depth=8} \\ \cline{5-16} 
\multicolumn{1}{c|}{} &  &  &  & CART & LS-OCT & DEOCT & MH-DEOCT & \begin{tabular}[c]{@{}c@{}}DEOCT\\      (Long)\end{tabular} & \begin{tabular}[c]{@{}c@{}}MH-DEOCT\\      (Long)\end{tabular} & CART & LS-OCT & DEOCT & MH-DEOCT & \begin{tabular}[c]{@{}c@{}}DEOCT\\      (Long)\end{tabular} & \begin{tabular}[c]{@{}c@{}}MH-DEOCT\\      (Long)\end{tabular} \\ \hline
\endhead
\hline
\endfoot
\endlastfoot
\multicolumn{1}{c|}{Soybean-small} & 47 & 35 & 4 & 3.51 & 3.51 & 5.27 & 3.51 & 5.00 & 0.00 & 3.51 & 3.51 & 6.97 & 0.00 & 10.87 & 0.00 \\
\multicolumn{1}{c|}{Echocardiogram} & 61 & 11 & 2 & 3.23 & 3.23 & 3.23 & 4.37 & 6.73 & 6.04 & 3.23 & 3.23 & 9.68 & 7.91 & 19.06 & 8.15 \\
\multicolumn{1}{c|}{Hepatitis} & 80 & 19 & 2 & 6.99 & 6.99 & 5.68 & 4.86 & 7.00 & 4.59 & 6.69 & 6.69 & 3.16 & 2.04 & 3.54 & 5.70 \\
\multicolumn{1}{c|}{Fertility} & 100 & 9 & 2 & 6.65 & 5.80 & 6.65 & 8.32 & 8.80 & 9.39 & 7.00 & 10.84 & 7.00 & 8.39 & 7.48 & 12.46 \\
\multicolumn{1}{c|}{Acute-inflammations-1} & 120 & 6 & 2 & 2.11 & 2.11 & 0.00 & 0.00 & 0.00 & 0.00 & 2.11 & 2.11 & 0.00 & 0.00 & 0.00 & 0.00 \\
\multicolumn{1}{c|}{Acute-inflammations-2} & 120 & 6 & 2 & 0.00 & 0.00 & 0.00 & 0.00 & 0.00 & 0.00 & 0.00 & 0.00 & 0.00 & 0.00 & 0.00 & 0.00 \\
\multicolumn{1}{c|}{Hayes–roth} & 132 & 5 & 3 & 4.33 & 6.19 & 7.26 & 9.81 & 11.34 & 8.35 & 6.23 & 12.29 & 7.82 & 17.92 & 3.95 & 12.96 \\
\multicolumn{1}{c|}{Iris} & 150 & 4 & 3 & 3.10 & 4.20 & 3.73 & 3.73 & 3.54 & 2.83 & 3.10 & 3.10 & 2.15 & 2.72 & 2.35 & 2.20 \\
\hline
\end{longtable}
}

\subsection{Training time (s) on 65 datasets within 1 million samples}
{\tiny\tabcolsep=3pt
\begin{longtable}{cccc|cccccccc}
\caption{Depth=2 average training time (s) on 65 datasets with 10 random splits within 1 million samples} \\
\label{tbl:d2trainingtime}\\
\hline
\multicolumn{1}{c|}{\multirow{2}{*}{Dataset}} & \multirow{2}{*}{Sample} & \multirow{2}{*}{Feature} & \multirow{2}{*}{Class} & \multicolumn{8}{c}{Depth=2} \\ \cline{5-12} 
\multicolumn{1}{c|}{} &  &  &  & CART & LS-OCT & Quant-BnB & RS-OCT & DEOCT & MH-DEOCT & \begin{tabular}[c]{@{}c@{}}DEOCT\\      (Long)\end{tabular} & \begin{tabular}[c]{@{}c@{}}MH-DEOCT\\      (Long)\end{tabular} \\ \hline
\endfirsthead
\multicolumn{12}{c}%
{{\bfseries Table \thetable\ continued from previous page}} \\
\hline
\multicolumn{1}{c|}{\multirow{2}{*}{Dataset}} & \multirow{2}{*}{Sample} & \multirow{2}{*}{Feature} & \multirow{2}{*}{Class} & \multicolumn{8}{c}{Depth=2} \\ \cline{5-12} 
\multicolumn{1}{c|}{} &  &  &  & CART & LS-OCT & Quant-BnB & RS-OCT & DEOCT & MH-DEOCT & \begin{tabular}[c]{@{}c@{}}DEOCT\\      (Long)\end{tabular} & \begin{tabular}[c]{@{}c@{}}MH-DEOCT\\      (Long)\end{tabular} \\ \hline
\endhead
\hline
\endfoot
\endlastfoot
\multicolumn{1}{c|}{Soybean-small} & 47 & 35 & 4 & 0.00 & 1.33 & 0.13 & 13.96 & 0.58 & 1.17 & 7.15 & 14.32 \\
\multicolumn{1}{c|}{Echocardiogram} & 61 & 11 & 2 & 0.00 & 0.42 & 0.07 & 18.22 & 0.37 & 0.74 & 3.96 & 7.96 \\
\multicolumn{1}{c|}{Hepatitis} & 80 & 19 & 2 & 0.00 & 0.90 & 0.09 & 163.05 & 0.48 & 0.97 & 5.66 & 11.34 \\
\multicolumn{1}{c|}{Fertility} & 100 & 9 & 2 & 0.00 & 0.66 & 0.06 & 169.94 & 0.38 & 0.75 & 4.15 & 8.28 \\
\multicolumn{1}{c|}{Acute-inflammations-1} & 120 & 6 & 2 & 0.00 & 1.01 & 0.04 & 18.40 & 0.47 & 1.51 & 4.42 & 9.48 \\
\multicolumn{1}{c|}{Acute-inflammations-2} & 120 & 6 & 2 & 0.00 & 0.43 & 0.02 & 13.92 & 0.37 & 0.79 & 3.88 & 8.13 \\
\multicolumn{1}{c|}{Hayes–roth} & 132 & 5 & 3 & 0.00 & 0.43 & 0.01 & 180.25 & 0.38 & 0.75 & 4.14 & 8.30 \\
\multicolumn{1}{c|}{Iris} & 150 & 4 & 3 & 0.00 & 0.38 & 0.06 & 113.50 & 0.37 & 0.73 & 4.02 & 8.05 \\
\hline
\end{longtable}
\vskip -0.3in
$\textbf{*}$ RS-OCT with training time larger than 14400s is recorded as 14400s in \cref{tbl:d2trainingtime}.
}

{\tiny\tabcolsep=3pt
\begin{longtable}{cccc|cccccccc}
\caption{Depth=2 training time standard deviation (s) on 65 datasets with 10 random splits within 1 million samples} \\
\hline
\multicolumn{1}{c|}{\multirow{2}{*}{Dataset}} & \multirow{2}{*}{Sample} & \multirow{2}{*}{Feature} & \multirow{2}{*}{Class} & \multicolumn{8}{c}{Depth=2} \\ \cline{5-12} 
\multicolumn{1}{c|}{} &  &  &  & CART & LS-OCT & Quant-BnB & RS-OCT & DEOCT & MH-DEOCT & \begin{tabular}[c]{@{}c@{}}DEOCT\\      (Long)\end{tabular} & \begin{tabular}[c]{@{}c@{}}MH-DEOCT\\      (Long)\end{tabular} \\ \hline
\endfirsthead
\multicolumn{12}{c}%
{{\bfseries Table \thetable\ continued from previous page}} \\
\hline
\multicolumn{1}{c|}{\multirow{2}{*}{Dataset}} & \multirow{2}{*}{Sample} & \multirow{2}{*}{Feature} & \multirow{2}{*}{Class} & \multicolumn{8}{c}{Depth=2} \\ \cline{5-12} 
\multicolumn{1}{c|}{} &  &  &  & CART & LS-OCT & Quant-BnB & RS-OCT & DEOCT & MH-DEOCT & \begin{tabular}[c]{@{}c@{}}DEOCT\\      (Long)\end{tabular} & \begin{tabular}[c]{@{}c@{}}MH-DEOCT\\      (Long)\end{tabular} \\ \hline
\endhead
\hline
\endfoot
\endlastfoot
\multicolumn{1}{c|}{Soybean-small} & 47 & 35 & 4 & 0.00 & 0.02 & 0.01 & 14.72 & 0.02 & 0.03 & 0.22 & 0.42 \\
\multicolumn{1}{c|}{Echocardiogram} & 61 & 11 & 2 & 0.00 & 0.03 & 0.01 & 18.88 & 0.01 & 0.02 & 0.10 & 0.22 \\
\multicolumn{1}{c|}{Hepatitis} & 80 & 19 & 2 & 0.00 & 0.01 & 0.01 & 94.06 & 0.01 & 0.02 & 0.16 & 0.44 \\
\multicolumn{1}{c|}{Fertility} & 100 & 9 & 2 & 0.00 & 0.01 & 0.01 & 119.81 & 0.00 & 0.01 & 0.06 & 0.13 \\
\multicolumn{1}{c|}{Acute-inflammations-1} & 120 & 6 & 2 & 0.00 & 0.03 & 0.10 & 19.02 & 0.27 & 0.02 & 0.29 & 0.04 \\
\multicolumn{1}{c|}{Acute-inflammations-2} & 120 & 6 & 2 & 0.00 & 0.01 & 0.00 & 14.67 & 0.01 & 0.20 & 0.01 & 0.20 \\
\multicolumn{1}{c|}{Hayes–roth} & 132 & 5 & 3 & 0.00 & 0.01 & 0.00 & 57.16 & 0.01 & 0.02 & 0.07 & 0.14 \\
\multicolumn{1}{c|}{Iris} & 150 & 4 & 3 & 0.00 & 0.01 & 0.00 & 75.20 & 0.00 & 0.00 & 0.06 & 0.10 \\
\hline
\end{longtable}
}

{\tiny\tabcolsep=3pt
\begin{longtable}{cccc|ccccccc}
\caption{Depth=3 average training time (s) on 65 datasets with 10 random splits within 1 million samples} \\
\hline
\multicolumn{1}{c|}{\multirow{2}{*}{Dataset}} & \multirow{2}{*}{Sample} & \multirow{2}{*}{Feature} & \multirow{2}{*}{Class} & \multicolumn{7}{c}{Depth=3} \\ \cline{5-11} 
\multicolumn{1}{c|}{} &  &  &  & CART & LS-OCT & Quant-BnB & DEOCT & MH-DEOCT & \begin{tabular}[c]{@{}c@{}}DEOCT\\      (Long)\end{tabular} & \begin{tabular}[c]{@{}c@{}}MH-DEOCT\\      (Long)\end{tabular} \\ \hline
\endfirsthead
\multicolumn{11}{c}%
{{\bfseries Table \thetable\ continued from previous page}} \\
\hline
\multicolumn{1}{c|}{\multirow{2}{*}{Dataset}} & \multirow{2}{*}{Sample} & \multirow{2}{*}{Feature} & \multirow{2}{*}{Class} & \multicolumn{7}{c}{Depth=3} \\ \cline{5-11} 
\multicolumn{1}{c|}{} &  &  &  & CART & LS-OCT & Quant-BnB & DEOCT & MH-DEOCT & \begin{tabular}[c]{@{}c@{}}DEOCT\\      (Long)\end{tabular} & \begin{tabular}[c]{@{}c@{}}MH-DEOCT\\      (Long)\end{tabular} \\ \hline
\endhead
\hline
\endfoot
\endlastfoot
\multicolumn{1}{c|}{Soybean-small} & 47 & 35 & 4 & 0.00 & 2.56 & - & 0.85 & 2.86 & 11.47 & 37.16 \\
\multicolumn{1}{c|}{Echocardiogram} & 61 & 11 & 2 & 0.00 & 0.73 & 13.43 & 0.54 & 1.50 & 6.66 & 17.89 \\
\multicolumn{1}{c|}{Hepatitis} & 80 & 19 & 2 & 0.00 & 1.66 & 8.33 & 0.70 & 2.23 & 8.94 & 27.76 \\
\multicolumn{1}{c|}{Fertility} & 100 & 9 & 2 & 0.00 & 0.89 & 0.49 & 0.54 & 1.80 & 6.64 & 21.88 \\
\multicolumn{1}{c|}{Acute-inflammations-1} & 120 & 6 & 2 & 0.00 & 1.30 & 5.89 & 0.53 & 2.20 & 22.02 & 40.91 \\
\multicolumn{1}{c|}{Acute-inflammations-2} & 120 & 6 & 2 & 0.00 & 0.73 & 0.31 & 0.52 & 1.56 & 6.29 & 18.54 \\
\multicolumn{1}{c|}{Hayes–roth} & 132 & 5 & 3 & 0.00 & 0.78 & 0.31 & 0.52 & 1.74 & 6.57 & 21.60 \\
\multicolumn{1}{c|}{Iris} & 150 & 4 & 3 & 0.00 & 0.71 & 1.23 & 0.51 & 1.71 & 6.44 & 20.77 \\
\hline
\end{longtable}
}

{\tiny\tabcolsep=3pt
\begin{longtable}{cccc|ccccccc}
\caption{Depth=3 training time standard deviation (s) on 65 datasets with 10 random splits within 1 million samples} \\
\hline
\multirow{2}{*}{Dataset} & \multirow{2}{*}{Sample} & \multirow{2}{*}{Feature} & \multirow{2}{*}{Class} & \multicolumn{7}{c}{Depth=3} \\ \cline{5-11} 
 &  &  &  & CART & LS-OCT & Quant-BnB & DEOCT & MH-DEOCT & \begin{tabular}[c]{@{}c@{}}DEOCT\\      (Long)\end{tabular} & \begin{tabular}[c]{@{}c@{}}MH-DEOCT\\      (Long)\end{tabular} \\ \hline
\endfirsthead
\multicolumn{11}{c}%
{{\bfseries Table \thetable\ continued from previous page}} \\
\hline
\multirow{2}{*}{Dataset} & \multirow{2}{*}{Sample} & \multirow{2}{*}{Feature} & \multirow{2}{*}{Class} & \multicolumn{7}{c}{Depth=3} \\ \cline{5-11} 
 &  &  &  & CART & LS-OCT & Quant-BnB & DEOCT & MH-DEOCT & \begin{tabular}[c]{@{}c@{}}DEOCT\\      (Long)\end{tabular} & \begin{tabular}[c]{@{}c@{}}MH-DEOCT\\      (Long)\end{tabular} \\ \hline
\endhead
\hline
\endfoot
\endlastfoot
Soybean-small & 47 & 35 & 4 & 0.00 & 0.03 & - & 0.01 & 0.04 & 0.59 & 1.58 \\
Echocardiogram & 61 & 11 & 2 & 0.00 & 0.01 & 5.87 & 0.01 & 0.13 & 0.33 & 1.36 \\
Hepatitis & 80 & 19 & 2 & 0.00 & 0.02 & 3.59 & 0.01 & 0.17 & 0.31 & 2.27 \\
Fertility & 100 & 9 & 2 & 0.00 & 0.01 & 0.02 & 0.00 & 0.12 & 0.21 & 0.90 \\
Acute-inflammations-1 & 120 & 6 & 2 & 0.00 & 0.01 & 12.60 & 0.03 & 0.19 & 5.88 & 6.98 \\
Acute-inflammations-2 & 120 & 6 & 2 & 0.00 & 0.02 & 0.04 & 0.01 & 0.18 & 0.22 & 2.08 \\
Hayes–roth & 132 & 5 & 3 & 0.00 & 0.01 & 0.04 & 0.01 & 0.15 & 0.23 & 0.77 \\
Iris & 150 & 4 & 3 & 0.00 & 0.01 & 0.11 & 0.01 & 0.15 & 0.24 & 1.68 \\
\hline
\end{longtable}
}

{\tiny\tabcolsep=3pt
\setlength{\LTleft}{-20cm plus -1fill}
\setlength{\LTright}{\LTleft}

\begin{longtable}{cccc|cccccc|cccccc}
\caption{Depth=4,8 average training time (s) on 65 datasets with 10 random splits within 1 million samples} \\
\hline
\multicolumn{1}{c|}{\multirow{2}{*}{Dataset}} & \multirow{2}{*}{Sample} & \multirow{2}{*}{Feature} & \multirow{2}{*}{Class} & \multicolumn{6}{c|}{Depth=4} & \multicolumn{6}{c}{Depth=8} \\ \cline{5-16} 
\multicolumn{1}{c|}{} &  &  &  & CART & LS-OCT & DEOCT & MH-DEOCT & \begin{tabular}[c]{@{}c@{}}DEOCT\\      (Long)\end{tabular} & \begin{tabular}[c]{@{}c@{}}MH-DEOCT\\      (Long)\end{tabular} & CART & LS-OCT & DEOCT & MH-DEOCT & \begin{tabular}[c]{@{}c@{}}DEOCT\\      (Long)\end{tabular} & \begin{tabular}[c]{@{}c@{}}MH-DEOCT\\      (Long)\end{tabular} \\ \hline
\endfirsthead
\multicolumn{16}{c}%
{{\bfseries Table \thetable\ continued from previous page}} \\
\hline
\multicolumn{1}{c|}{\multirow{2}{*}{Dataset}} & \multirow{2}{*}{Sample} & \multirow{2}{*}{Feature} & \multirow{2}{*}{Class} & \multicolumn{6}{c|}{Depth=4} & \multicolumn{6}{c}{Depth=8} \\ \cline{5-16} 
\multicolumn{1}{c|}{} &  &  &  & CART & LS-OCT & DEOCT & MH-DEOCT & \begin{tabular}[c]{@{}c@{}}DEOCT\\      (Long)\end{tabular} & \begin{tabular}[c]{@{}c@{}}MH-DEOCT\\      (Long)\end{tabular} & CART & LS-OCT & DEOCT & MH-DEOCT & \begin{tabular}[c]{@{}c@{}}DEOCT\\      (Long)\end{tabular} & \begin{tabular}[c]{@{}c@{}}MH-DEOCT\\      (Long)\end{tabular} \\ \hline
\endhead
\hline
\endfoot
\endlastfoot
\multicolumn{1}{c|}{Soybean-small} & 47 & 35 & 4 & 0.00 & 3.41 & 1.25 & 5.24 & 16.13 & 67.17 & 0.00 & 32.19 & 11.17 & 27.57 & 115.10 & 310.61 \\
\multicolumn{1}{c|}{Echocardiogram} & 61 & 11 & 2 & 0.00 & 1.20 & 0.82 & 2.68 & 10.64 & 32.77 & 0.00 & 13.06 & 8.10 & 19.11 & 96.74 & 207.36 \\
\multicolumn{1}{c|}{Hepatitis} & 80 & 19 & 2 & 0.00 & 2.43 & 1.01 & 4.20 & 13.15 & 54.07 & 0.00 & 25.56 & 7.32 & 19.76 & 90.78 & 266.78 \\
\multicolumn{1}{c|}{Fertility} & 100 & 9 & 2 & 0.00 & 1.61 & 0.82 & 3.51 & 10.29 & 44.63 & 0.00 & 17.39 & 7.15 & 21.82 & 81.86 & 282.83 \\
\multicolumn{1}{c|}{Acute-inflammations-1} & 120 & 6 & 2 & 0.00 & 1.72 & 0.82 & 3.29 & 24.83 & 53.59 & 0.00 & 14.02 & 11.42 & 19.58 & 108.14 & 176.26 \\
\multicolumn{1}{c|}{Acute-inflammations-2} & 120 & 6 & 2 & 0.00 & 1.15 & 0.80 & 2.72 & 9.90 & 32.45 & 0.00 & 12.62 & 7.19 & 12.53 & 93.14 & 157.59 \\
\multicolumn{1}{c|}{Hayes–roth} & 132 & 5 & 3 & 0.00 & 1.34 & 0.79 & 3.53 & 9.93 & 45.68 & 0.00 & 16.29 & 6.75 & 29.46 & 55.19 & 356.30 \\
\multicolumn{1}{c|}{Iris} & 150 & 4 & 3 & 0.00 & 1.05 & 0.78 & 3.42 & 9.55 & 39.78 & 0.00 & 15.77 & 7.67 & 28.13 & 56.92 & 203.80 \\
\hline
\end{longtable}
}

{\tiny\tabcolsep=3pt
\setlength{\LTleft}{-20cm plus -1fill}
\setlength{\LTright}{\LTleft}
\begin{longtable}{cccc|cccccc|cccccc}
\caption{Depth=4,8 training time standard deviation (s) on 65 datasets with 10 random splits within 1 million samples} \\
\hline
\multicolumn{1}{c|}{\multirow{2}{*}{Dataset}} & \multirow{2}{*}{Sample} & \multirow{2}{*}{Feature} & \multirow{2}{*}{Class} & \multicolumn{6}{c|}{Depth=4} & \multicolumn{6}{c}{Depth=8} \\ \cline{5-16} 
\multicolumn{1}{c|}{} &  &  &  & CART & LS-OCT & DEOCT & MH-DEOCT & \begin{tabular}[c]{@{}c@{}}DEOCT\\      (Long)\end{tabular} & \begin{tabular}[c]{@{}c@{}}MH-DEOCT\\      (Long)\end{tabular} & CART & LS-OCT & DEOCT & MH-DEOCT & \begin{tabular}[c]{@{}c@{}}DEOCT\\      (Long)\end{tabular} & \begin{tabular}[c]{@{}c@{}}MH-DEOCT\\      (Long)\end{tabular} \\ \hline
\endfirsthead
\multicolumn{16}{c}%
{{\bfseries Table \thetable\ continued from previous page}} \\
\hline
\multicolumn{1}{c|}{\multirow{2}{*}{Dataset}} & \multirow{2}{*}{Sample} & \multirow{2}{*}{Feature} & \multirow{2}{*}{Class} & \multicolumn{6}{c|}{Depth=4} & \multicolumn{6}{c}{Depth=8} \\ \cline{5-16} 
\multicolumn{1}{c|}{} &  &  &  & CART & LS-OCT & DEOCT & MH-DEOCT & \begin{tabular}[c]{@{}c@{}}DEOCT\\      (Long)\end{tabular} & \begin{tabular}[c]{@{}c@{}}MH-DEOCT\\      (Long)\end{tabular} & CART & LS-OCT & DEOCT & MH-DEOCT & \begin{tabular}[c]{@{}c@{}}DEOCT\\      (Long)\end{tabular} & \begin{tabular}[c]{@{}c@{}}MH-DEOCT\\      (Long)\end{tabular} \\ \hline
\endhead
\hline
\endfoot
\endlastfoot
\multicolumn{1}{c|}{Soybean-small} & 47 & 35 & 4 & 0.00 & 0.05 & 0.04 & 0.42 & 5.17 & 8.52 & 0.00 & 2.73 & 0.27 & 3.34 & 9.82 & 25.97 \\
\multicolumn{1}{c|}{Echocardiogram} & 61 & 11 & 2 & 0.00 & 0.02 & 0.01 & 0.58 & 0.20 & 3.75 & 0.00 & 0.31 & 0.28 & 3.32 & 2.72 & 15.50 \\
\multicolumn{1}{c|}{Hepatitis} & 80 & 19 & 2 & 0.00 & 0.05 & 0.24 & 0.56 & 0.13 & 0.43 & 0.00 & 1.82 & 0.75 & 4.27 & 1.27 & 38.03 \\
\multicolumn{1}{c|}{Fertility} & 100 & 9 & 2 & 0.00 & 0.14 & 0.01 & 0.03 & 0.63 & 0.94 & 0.00 & 3.17 & 0.23 & 5.45 & 0.65 & 65.74 \\
\multicolumn{1}{c|}{Acute-inflammations-1} & 120 & 6 & 2 & 0.00 & 0.11 & 0.02 & 0.02 & 0.18 & 0.49 & 0.00 & 2.21 & 0.09 & 3.34 & 4.49 & 75.42 \\
\multicolumn{1}{c|}{Acute-inflammations-2} & 120 & 6 & 2 & 0.00 & 0.78 & 0.02 & 0.27 & 0.66 & 1.55 & 0.00 & 10.12 & 0.09 & 7.18 & 4.61 & 95.39 \\
\multicolumn{1}{c|}{Hayes–roth} & 132 & 5 & 3 & 0.00 & 0.06 & 0.03 & 0.31 & 0.59 & 4.30 & 0.00 & 2.10 & 0.22 & 6.92 & 4.13 & 95.29 \\
\multicolumn{1}{c|}{Iris} & 150 & 4 & 3 & 0.00 & 0.16 & 0.01 & 0.12 & 0.43 & 1.08 & 0.00 & 7.88 & 1.02 & 4.58 & 2.14 & 42.47 \\
\hline
\end{longtable}
}